%% file: aistats2026.tex
\documentclass[twoside]{article}
\usepackage[accepted]{aistats2026}

\usepackage[utf8]{inputenc} 
\usepackage[T1]{fontenc}    
\usepackage[colorlinks=true,urlcolor=black,citecolor=blue,menucolor=black,linkcolor=blue]{hyperref}       
\usepackage{url}            
\usepackage{booktabs}       
\usepackage{amsfonts}       
\usepackage{nicefrac}       
\usepackage{microtype}      
\usepackage{xcolor}         
\usepackage{amsmath,amsfonts,amsthm,bm} %
\usepackage{cleveref}
\usepackage{thm-restate}
\usepackage{graphicx}
\usepackage{algorithm} 
\usepackage[noend]{algpseudocode} 
\usepackage{multicol}
\usepackage{multirow}
\usepackage[percent]{overpic}
\usepackage{caption}
\usepackage{subcaption}
\newtheorem{theorem}{Theorem}
\newtheorem{proposition}{Proposition}
\newtheorem{lemma}{Lemma}

\newtheorem{definition}{Definition}
\theoremstyle{remark}

\usepackage[round]{natbib}

\input{./commands.tex}

\definecolor{ChangeColor}{HTML}{ff0000}

\begin{document}

%
\runningtitle{Probabilistic Cuts for Differentiable Graph Partitioning}



\twocolumn[
  \aistatstitle{Beyond Spectral Clustering: Probabilistic Cuts for Differentiable Graph Partitioning}
  \aistatsauthor{Ayoub Ghriss}
  \aistatsaddress{\texttt{ayoub.ghriss@polytechnique.org}\\ Department of Computer Science\\ University of Colorado, Boulder}
]

\begin{abstract}
  Probabilistic relaxations of graph cuts offer a differentiable alternative to spectral
  clustering, enabling end-to-end and online learning without eigendecompositions, yet
  prior work centered on RatioCut and lacked general guarantees and principled gradients.
  We present a unified probabilistic framework that covers a wide class of cuts, including
  Normalized Cut. Our framework provides tight analytic upper bounds on expected discrete
  cuts via integral representations and Gauss hypergeometric functions with closed-form
  forward and backward. Together, these results deliver a rigorous, numerically stable
  foundation\footnote{\href{https://github.com/ayghri/pgcuts}{\color{purple} Github: \texttt{https://github.com/ayghri/pgcuts}}} for scalable,
  differentiable graph partitioning covering a wide range of
  clustering and contrastive learning objectives.
\end{abstract}

\section{Introduction}\label{sec:introduction}
Self-Supervised Learning (SSL) has become the backbone of modern representation learning, closing
the gap to supervised baselines across vision, speech, and
language~\citep{radford21clip,baevski20wav2vec2}. Successful objectives broadly fall into two
families: \emph{contrastive} methods that optimize pairwise
relationships~\citep{chen20simclr,oord18cpc,he20moco}, and \emph{non-contrastive} or \emph{masked}
approaches that rely on local reconstruction or
invariance~\citep{grill20byol,simeoni25dinov3,he22mae}. While effective, clustering-based variants
such as DeepCluster~\citep{caron18deepcluster} or SwAV~\citep{caron20swav} often impose Voronoi
tessellations on the latent space. These assumptions bias models toward convex, globular clusters,
failing to capture the intrinsic \emph{manifold structure} of complex data distributions.

The ideal alternative can be achieved through graph cut-based partitioning, more specifically
through the \emph{Normalized Cut} (NCut)~\citep{shi00ncut}. Unlike simple Ratio Cuts (RCut); which
balance partitions based on node counts; NCut balances partitions based on \emph{volume} (total
edge weight). This defines distance not by Euclidean proximity, but by connectivity through
\emph{dense domains}, approximating geodesic distance on the underlying manifold. Despite this
geometric superiority, NCut remains largely incompatible with modern deep learning: standard
spectral relaxations require solving generalized eigenvalue problems, a process that is
computationally expensive ($O(N^3)$ in dense setting) and notoriously unstable to differentiate
end-to-end.

In this work, we bridge the divide between the geometric purity of graph cuts and the scalability
of modern SSL. We present a \emph{unified, differentiable probabilistic framework} that relaxes the
discrete graph cut problem without resorting to spectral decomposition by extending prior work
of~\citet{prcut}. By treating cluster assignments as probabilistic variables, we resolve the
challenge of evaluating the expected volume-normalized cut; an expectation of a ratio using
\emph{Gauss hypergeometric functions}. This yields tight, analytic upper bounds that turn the
discrete cut-based objectives into a smooth, stable surrogate.

Our contributions are as follows:
\begin{itemize}
  \item We derive a probabilistic relaxation for a wide class of graph cuts, including Normalized Cut.
        Unlike prior approximations, our framework respects the volume constraints required for
        manifold-aware partitioning.
  \item We provide closed-form forward and backward passes using hypergeometric polynomials, establishing a
        numerically stable foundation for scalable differentiation that avoids eigendecompositions
        entirely.
  \item We rigorously control the approximation error via two-sided AM–GM gap bounds and a zero-aware
        penalty, ensuring the surrogate faithfully minimizes the true expected cut.
  \item We connect this geometric view back to SSL, showing that widely used contrastive objectives (e.g.,
        SimCLR, CLIP) emerge as special cases of our envelope when the graph is constructed from batch
        embeddings.
\end{itemize}

\section{Preliminaries} \label{sec:prelimanaries}
Let $\gG=(V,E,\mW)$ be an undirected weighted graph on $n=|V|$ vertices with a symmetric,
elementwise nonnegative adjacency matrix $\mW\in\R^{n\times n}$; assume $\mW_{ii}=0$. Define the
degree of $i\in V$ by $d_i \defeq \sum_{v_j\in V}\mW_{ij}$ and the degree matrix $\mD \defeq
  \diag(d_1,\ldots,d_n)$.

\begin{figure}
  \begin{center}
    \begin{overpic}[width=0.4\textwidth]{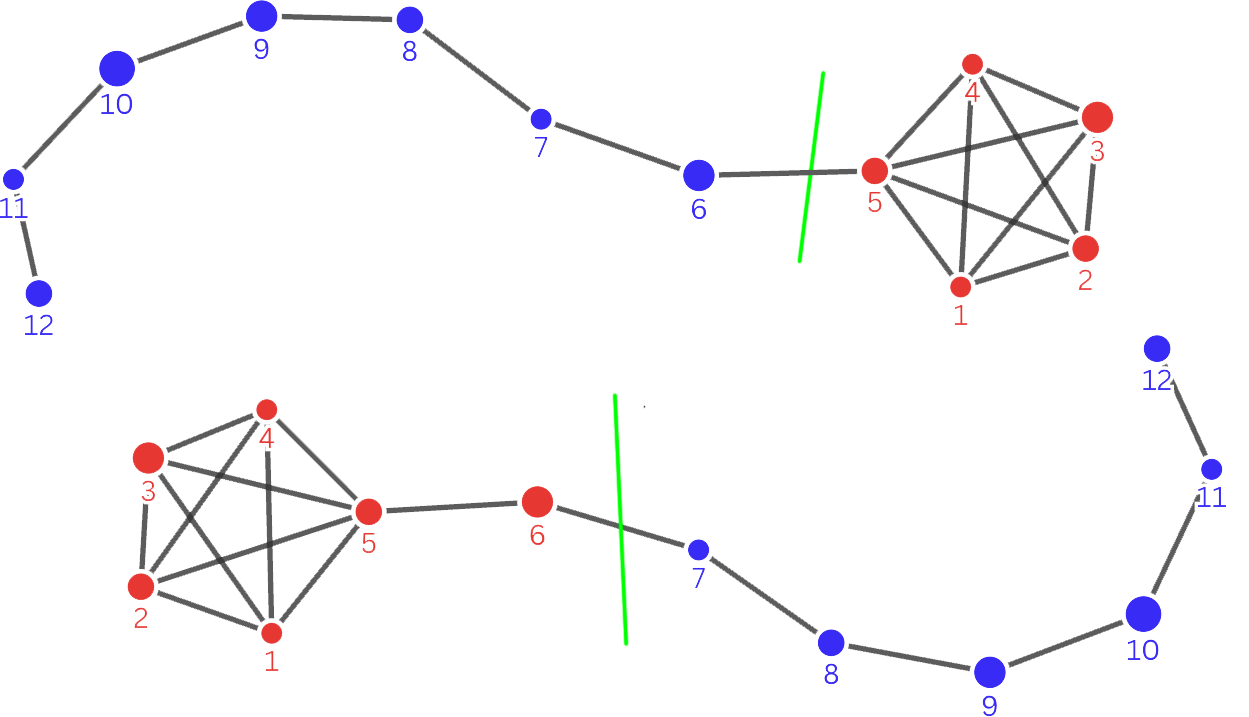}
      \put(50,50){Kite $1$}
      \put(25,40){\textbf{Best N-Cut}}
      \put(30,22){Kite $2$}
      \put(25,3){\textbf{Best R-Cut}}
    \end{overpic}
  \end{center}
  \caption{
    In the kite graph with $\emW_{i,j}=1$ for connected nodes,
    RCut maximizes the average within-cluster edge weight,
    while NCut identifies clusters with densely connected nodes.
    Using~\Cref{eq:generic-cut} on Kite $1$ and Kite $2$, NCut evaluates to, respectively,
    $\lrbp{\frac{33}{260},\frac{33}{242}}$, while RCut evaluates to
    $\lrbp{\frac{12}{35}, \frac{12}{36}}$.
  }
  \label{fig:rcut_vs_ncut}
\end{figure}

For $A\subseteq V$, let $\comp{A}\defeq V\setminus A$ and identify $A$ by its indicator vector
$\vone_{A}\in\{0,1\}^n$. The cut associated with $A$ is:
\begin{align*}
  \cut(A)=\cut(\comp{A})\defeq \sum_{(v_i,v_j)\in A\times\comp{A}} \mW_{ij}
  \;=\; \vone_{A}^\top \mW\,\vone_{\comp{A}},
\end{align*}
and the associated volume-normalized cut is:
\begin{equation}
  \label{eq:generic-cut}
  \vcut(A)\;\defeq\; \frac{\cut(A)}{\volm(A)},
\end{equation}
where the \emph{volume} is $\volm(A)\defeq \sum_{v_i\in A} s(v_i)$ for a given vertex
\textit{size} function $s: V\to\R_{>0}$. For example, the ratio cut (\textbf{RCut})
uses $s(v_i)\equiv 1$ so $\volm\lrp{A}\equiv|A|$, whereas the normalized cut
(\textbf{NCut}) uses $s(v_i)=d_i$. This difference can yield different partitioning that minimizes
the volume-normalized cuts as shown in~\Cref{fig:rcut_vs_ncut}.

We fix the size function $s$ and write $\vs\defeq(s_1,\cdots, s_n)$ with $s_i\defeq s(v_i)$. The
goal is to find a $k$-way clustering $\gC_k=\{\sC_\ell\}_{\ell=1}^k$ of $V$ that minimizes the
volume-normalized graph cut:
\begin{equation}\label{eq:gcut}
  \gcut(\gC_k)\;\defeq\; \frac{1}{2}\sum_{\ell=1}^{k}\vcut(\sC_\ell).
\end{equation}

\subsection{Probabilistic Relaxation}
The Probabilistic Ratio-Cut (PRCut)~\citep{prcut} adopts a probabilistic relaxation of $k$-way
clustering. Let $\rva_{\ell}\in\{0,1\}^n$ be the random indicator of the cluster $\sC_\ell$. The
clustering $\gC_k$ is parameterized by a row-stochastic matrix $\mP\in[0,1]^{n\times k}$ with
$\sum_{\ell=1}^k \mP_{i\ell}=1$ and $\mP_{i\ell}=\Prob{\erva_{\ell,i}=1}=\Prob{v_i\in\sC_\ell}$.

The expected graph cut is defined as:
\begin{equation}\label{eq:gcut_prob}
  \gcut(\mP)\;\defeq\; \frac{1}{2}\sum_{\ell=1}^{k}\Ea{\wh{\vcut}(\rva_\ell)},
\end{equation}
where
\begin{equation}
  \wh{\vcut}(\rva_\ell)\;\defeq\;
  \frac{\rva_\ell^\top \mW\,(\vone-\rva_\ell)}{\vs^\top \rva_\ell}.
\end{equation}
The following bound underpins the PRCut framework:
\begin{proposition}[PRCut bound~\citep{prcut}]\label{prop:prcut_upper}
  For the ratio cut ($s_i\equiv 1,\; \forall i\in\{1,\ldots,n\}$):
  \[
    \Ea{\wh{\vcut}_R(\rva_\ell)}
    \le
    \frac{1}{n\,\ov{\mP_{:,\ell}}}
    \sum_{i,j=1}^n
    \emW_{ij}\bigl(\emP_{i\ell}+\emP_{j\ell}-2\,\emP_{i\ell}\emP_{j\ell}\bigr),
  \]
  where $\ov{\mP_{:,\ell}} \defeq \tfrac{1}{n}\sum_{i=1}^n \emP_{i\ell}$ denotes the expected
  fraction of vertices assigned to $\sC_\ell$.
\end{proposition}

In this paper, we derive tighter, more general bounds for an arbitrary vertex-weight function $s$,
with concentration guarantees and gradients that are compatible with first-order optimization.
\section{Proposed Method}
\input{./method.tex}
\section{Related Work}
\input{./related.tex}

\input{./results.tex}
\section{Conclusion}

We presented a probabilistic framework for differentiable graph partitioning that provides tight
upper bounds on the expected Normalized Cut via the Gauss hypergeometric function
${}_2F_1(-m,b;c;z)$. Our approach extends prior work~\citep{prcut} in three ways: (1)~we derive a
tighter hypergeometric envelope for NCut and RatioCut, handling heterogeneous vertex degrees
through a H\"older-product binning scheme; (2)~we introduce a gradient mixing strategy that
prevents cluster collapse without tuning loss weights; and (3)~we provide full implementation of
various methods with GPU-accelerated ${}_2F_1$ kernels (Triton and CUDA) with closed-form forward
and backward passes, making the bound practical for first-order optimization and stochastic gradient.

Experiments on 15 datasets with three foundation model embeddings show that: the parametric H-Cut
objectives consistently outperform non-parametric spectral clustering in clustering accuracy, while
spectral clustering achieves lower raw cut values---the linear model acts as an implicit
regularizer. The graph quality metric $Q$ reliably predicts when graph-cut methods will succeed,
confirming that the similarity graph is the primary bottleneck.

\paragraph{Future directions.}
Our framework takes a fixed similarity graph as input and optimizes cluster assignments. A natural
extension is to \emph{learn the similarity and clustering jointly}, closing the loop between graph
construction and cut optimization. This would enable learning embeddings that are linearly
separable by design. The parameterization from embeddings to cluster assignments (e.g., linear vs.\
nonlinear, shared vs.\ per-cluster) directly shapes the embedding space geometry. Such end-to-end
training could connect differentiable cuts to metric learning and structured representation
learning, where the graph encodes the desired invariances.

\paragraph{Limitations.}
Our analysis assumes independent Bernoulli assignments; modeling assignment dependencies (e.g., MRF
couplings) remains open. The H\"older envelope is tightest when the degree distribution within each
bin is concentrated; extremely heavy-tailed graphs may require adaptive binning. The ${}_2F_1$
evaluation is stable for small $a=-m$ (finite polynomial), but very large $m$ might benefit from
compensated summation.

\bibliography{aistats2026}

\clearpage
\appendix
\thispagestyle{empty}
\onecolumn
\aistatstitle{Appendices}
\input{./appendix.tex}

\end{document}

%% file: commands.tex
\DeclareMathOperator*{\cut}{cut}
\DeclareMathOperator*{\volm}{vol}
\newcommand{\gcut}{\text{GraphCut}}
\DeclareMathOperator*{\rcut}{RatioCut}
\DeclareMathOperator*{\vcut}{VolCut}
\DeclareMathOperator*{\ncut}{NCut}

\newcommand{\Ea}[1]{\E\left[#1\right]}

\newcommand{\Prob}[1]{\operatorname{Pr}\lrp{#1}}
\newcommand{\by}[1]{{\frac{1}{#1}}}

\newcommand{\comp}[1]{\overline{#1}}

\newcommand{\defeq}{\stackrel{\text{def}}{=}}
\newcommand{\diag}{\mathop{\mathrm{diag}}}
\newcommand{\hyperg}{{}_2F_1}
\newcommand{\kl}[2]{D_{\mathrm{KL}}(#1 \ \| \ #2)}

\newcommand{\lrcb}[1]{\left\lbrace#1\right\rbrace}
\newcommand{\lrp}[1]{\left(#1\right)}
\newcommand{\lrbp}[1]{\big(#1\big)}

\newcommand{\one}[1]{{1}_{#1}}
\newcommand{\ov}[1]{\overline{#1}}

\newcommand{\wh}[1]{\widehat{#1}}

\def\eqref#1{equation~\ref{#1}}

\def\1#1{\bm{1}_{#1}}

\def\rr{{\textnormal{r}}}

\def\rx{{\textnormal{x}}}

\def\rva{{\mathbf{a}}}

\def\erva{{\textnormal{a}}}

\def\vone{{\bm{1}}}

\def\valpha{{\bm{\alpha}}}
\def\vbeta{{\bm{\beta}}}

\def\vp{{\bm{p}}}

\def\vs{{\bm{s}}}

\def\vx{{\bm{x}}}

\def\vz{{\bm{z}}}

\def\evs{{s}}

\def\mD{{\bm{D}}}

\def\mL{{\bm{L}}}

\def\mP{{\bm{P}}}

\def\mT{{\bm{T}}}

\def\mW{{\bm{W}}}

\DeclareMathAlphabet{\mathsfit}{\encodingdefault}{\sfdefault}{m}{sl}
\SetMathAlphabet{\mathsfit}{bold}{\encodingdefault}{\sfdefault}{bx}{n}

\def\gA{{\mathcal{A}}}

\def\gC{{\mathcal{C}}}

\def\gE{{\mathcal{E}}}

\def\gG{{\mathcal{G}}}
\def\gH{{\mathcal{H}}}
\def\gI{{\mathcal{I}}}

\def\gV{{\mathcal{V}}}

\def\cL{{\mathcal{L}}}
\def\cB{{\mathcal{B}}}

\def\sA{{\mathbb{A}}}

\def\sC{{\mathbb{C}}}

\def\sA{{\mathbb{A}}}

\def\sJ{{\mathbb{J}}}

\def\emP{{P}}

\def\emW{{W}}

\newcommand{\E}{\mathbb{E}}

\newcommand{\R}{\mathbb{R}}

\newcommand{\softmax}{\mathrm{softmax}}

\newcommand{\Var}{\mathrm{Var}}



%% file: method.tex
By symmetry in~\Cref{eq:gcut_prob}, it suffices to bound the expected $\vcut$ for a single cluster.
Fix a cluster $\sC$ and drop the index $\ell$. Let $\rva\in\{0,1\}^n$ be its random indicator with
independent coordinates $\erva_i\sim\mathrm{Bernoulli}(p_i)$, and let
$\vp=(p_1,\ldots,p_n)^\top\in[0,1]^n$.
\begin{align}
  \wh{\vcut}(\rva) \;=\; \sum_{i,j=1}^n \mW_{ij}\,
  \frac{\erva_i(1-\erva_j)}{\sum_{l=1}^n \evs_l\,\erva_l}.
\end{align}
Without loss of generality, let $(i,j)=(1,2)$(note that $\mW_{11}=\mW_{22}=0$) and write:
\begin{align*}
  \Ea{\frac{\erva_1(1-{\erva_2})}{\sum_{i=1}^n s_i\,\erva_i}}
  \;=\;
  p_1(1-p_2)\;
  \Ea{\frac{1}{\,s_1 + \sum_{i=3}^n s_i\,\erva_i}\,}.
\end{align*}
Thus, we must evaluate expectations of the form $\Ea{1/(q+\rx)}$ with $q>0$ and
$\rx=\sum_{i=3}^n s_i\,\erva_i$. It turns out that $\rx$ follows a generalized
Poisson–Binomial distribution.
The Poisson–Binomial distribution is well studied and has applications across seemingly
unrelated areas~\citep{chen97poissonbinomial,cam60poissonbinomial}. We use its
generalized form:
\begin{definition}[Generalized Poisson–Binomial (GPB)]
  Let $\valpha\in[0,1]^m$ and $\theta_i<\beta_i$ be real constants, and let
  $\rr_i \sim \mathrm{Bernoulli}(\alpha_i)$ independently.
  The random variable $\rx \;=\; \sum_{i=1}^m \lrbp{\theta_i(1-\rr_i)+\beta_i \rr_i}$
  follows a generalized Poisson–Binomial distribution~\citep{zhang17gpb}.
\end{definition}
In our setting, $m\defeq n-2$, $\valpha=(p_3,\ldots,p_n)$, and the weights are
$(\theta_i,\beta_i)=(0,s_i)$, so $\rx=\sum_{i=3}^n s_i\,\rr_i$. We denote this special case
by $\operatorname{GPB}(\valpha,\vbeta)$, and compute its probability generating function (PGF) $G_\rx$:
\begin{align}\label{eq:gpb_pgf}
  G_\rx(t) \defeq\Ea{t^{\rx}}
  = \prod_{i=1}^m \lrbp{1-\alpha_i+\alpha_i\,t^{\,\beta_i}}, \; t\in[0,1].
\end{align}
The target expectation can now be computed via the identity $x^{-1}=\int_0^1
  t^{x-1}\,dt$ for $x>0$ (see~\Cref{appendix:integralexp}):
\begin{restatable}[Integral representation]{lemma}{integralexp}
  \label{lemma:integralexp}
  Define the integral $\gI(q,\valpha,\vbeta)$ as:
  \begin{equation}\label{eq:integral-def}
    \gI(q,\valpha,\vbeta) \;\defeq\;
    \int_0^1 t^{\,q-1}\, \prod_{i=1}^m \lrbp{1-\alpha_i+\alpha_i\,t^{\,\beta_i}}\,dt.
  \end{equation}
  For any $q>0$, we have:
  \begin{equation}\label{eq:reciprocal-int}
    \Ea{\frac{1}{q+\rx}} \;=\;\gI(q,\valpha,\vbeta)
  \end{equation}
\end{restatable}
For the ratio cut, $q=1$ and $\beta_i\equiv 1$, and PRCut uses the bound
$\Ea{\tfrac{1}{1+\rx}} \le (\sum_i \alpha_i)^{-1}$. In this work, $s$ need not be
constant, so different tools are required.

We first consider the case $\beta_i\equiv \beta$ and recall Gauss’s hypergeometric function
${}_2F_1$~\citep{chambers92hyperg}, defined for $|z|<1$ by the absolutely convergent power series:
\begin{equation}\label{eq:2F1-series}
  \hyperg(a,b;c;z) \;=\; \sum_{k=0}^{\infty}\frac{(a)_k\,(b)_k}{(c)_k}\,\frac{z^k}{k!},
\end{equation}
where $(x)_k \defeq x(x+1)\cdots(x+k-1)$ is the rising factorial, with $(x)_0\defeq 1$.
\begin{lemma}[Euler’s identity]
  If $c>b>0$ and $z\in[0,1]$, then ${}_2F_{1}(a,b;c;z)$ is equal to:
  \begin{equation}\label{eq:Euler-integral}
    \frac{\Gamma(c)}{\Gamma(b)\,\Gamma(c-b)}
    \int_{0}^{1} t^{\,b-1}(1-t)^{\,c-b-1}\,(1-zt)^{-a}\,dt,
  \end{equation}
  where $\Gamma$ denotes the gamma function.
\end{lemma}
A useful property of $\hyperg$ is the derivative formula:
\begin{equation}\label{eq:2F1-derivative}
  \frac{d}{dz}\,\hyperg(a,b;c;z) \;=\; \frac{ab}{c}\,{}_2F_{1}(a+1,b+1;c+1;z),
\end{equation}
which, in particular, implies the following:
\begin{lemma}[Properties of ${}_2F_1$]\label{lem:2F1-lip}
  Let $m\in\mathbb{N}$, $b>0$, and $c>b$. On $[0,1]$, the function
  $f(z)\defeq{}_2F_1(-m,b;c;z)$ is a degree-$m$ polynomial that is decreasing, convex,
  and $L$-Lipschitz with $L=\tfrac{mb}{c}$.
\end{lemma}
The integral in~\Cref{lemma:integralexp} admits a computable
and differentiable upper bound (proof in~\Cref{appendix:hyperbound}).
\begin{restatable}[Hypergeometric bound]{theorem}{hyperbound}\label{thm:hyperbound}
  Assume $\beta_i\equiv\beta>0$. For any $q>0$,
  \begin{equation}\label{eq:hyperbound}
    \gI(q,\valpha,\vbeta)
    \;\le\; \gH_\beta(q;\bar \alpha,m)
  \end{equation}
  where  $\gH_\beta(q;\bar \alpha,m) \defeq \frac{1}{q}\;
    \hyperg\!\lrp{-m,\,1;\,\frac{q}{\beta}+1;\,\bar\alpha}$, and $\bar \alpha \defeq \by{m}\sum_{i=1}^m \alpha_i$.
\end{restatable}

\subsection{The Envelope Gap} \label{sec:amgm-gap}
To quantify the tightness of the bound from~\Cref{thm:hyperbound}, we derive a pointwise Arithmetic
Mean-Geometric Mean (\textbf{AM-GM}) gap.
\begin{restatable}[Integrated AM--GM gap]{proposition}{amgmgap}\label{prop:amgm_gap}
  Let $\beta_i\equiv\beta>0$ and $\valpha\in[0,1]^m$. Define $h(t)=t^{q-1}\lrp{1-\bar \alpha+\bar \alpha\,t^{\beta}}^m$
  and:
  \begin{align*}
    \underline{\Delta}(q,\valpha)
     & \defeq \int_{0}^{1} h(t) \Bigl(1-e^{-\gamma(t)\Var(\valpha)}\Bigr)\,dt, \\
    \overline{\Delta}(q,\valpha)
     & \defeq \int_{0}^{1} h(t) \Bigl(1-e^{-\theta(t)\Var(\valpha)}\Bigr)\,dt,
  \end{align*}
  with $\gamma(t)\defeq\tfrac{m}{2}(1-t^{\beta})^2$ and $\theta(t)\defeq \gamma(t)/t^{2\beta}$.

  We have the following gap:
  \begin{align}
    \underline{\Delta}(q,\valpha)
    \le
    \gH_\beta(q;\bar\alpha,m)\;-\;\gI(q,\valpha,\vbeta)
    \le
    \overline{\Delta}(q,\valpha),
  \end{align}
  with $\Var(\valpha)$ computed under uniform sampling of the graph nodes, and
  equality throughout iff $\Var(\valpha)=0$.
\end{restatable}
A convenient corollary gives an explicit upper bound.
\begin{restatable}[Simple upper bound]{corollary}{simple_amgm}\label{cor:simple_amgm}
  Under the conditions of~\Cref{prop:amgm_gap}, for any $q>0$,
  \[
    \gH_\beta(q;\bar\alpha,m)-\gI(q,\valpha,\vbeta)
    \le
    \frac{m}{2}\Var(\valpha)\int_{0}^{1} h(t)\theta(t)\,dt.
  \]
\end{restatable}
See~\Cref{appendix:amgm_gap} for the proofs of~\Cref{prop:amgm_gap} and its corollary.
\paragraph{Zero-aware gap control.}
If we examine~\Cref{eq:integral-def}, we see that coordinates with $\alpha_i=0$ contribute the
factor $(1-\alpha_i+\alpha_i t^{\beta})\equiv 1$ and thus have no influence on
$\gI(q,\valpha,\vbeta)$. The AM--GM gap in~\Cref{prop:amgm_gap} over-penalizes configurations with
many inactive entries: zeros still inflate $\Var(\valpha)$ even though they do not affect the
product inside the integral. We therefore replace the plain variance by a \emph{zero-aware}
weighted dispersion that vanishes at $\alpha_i=0$. Let $\omega_0(x)\defeq x$ (more generally,
$\omega_0(x)=x^a, a\in[1,2]$), and define:
\begin{align}
  \Omega \;\defeq\; \sum_{i=1}^m \omega_0(\alpha_i),\qquad
  \bar\alpha^{\omega_0} \;\defeq\; \frac{1}{\Omega}\sum_{i=1}^m \omega_0(\alpha_i)\,\alpha_i, \\
  \Var^{\omega_0}(\valpha) \;\defeq\;
  \frac{1}{\Omega}\sum_{i=1}^m \omega_0(\alpha_i)\bigl(\alpha_i-\bar\alpha^{\omega_0}\bigr)^2
\end{align}

\begin{definition}
  Let $\gH_\beta(q;\bar\alpha,m)$ be the envelope from~\Cref{thm:hyperbound}. Define the second
  forward $\beta$-difference:
  \begin{align}
    \Delta(q;\bar\alpha,\beta,m) \defeq \sum_{r=0}^{2}\binom{2}{r}(-1)^r\;\gH_\beta\!\bigl(q+r\beta;\bar\alpha,m\bigr).
  \end{align}
  and $\gA (q;\bar \alpha,\beta,m) = \frac{\partial}{\partial \bar \alpha}
    \Delta(q;\bar\alpha,\beta,m)$.
\end{definition}

\begin{restatable}[Zero-aware AM-GM gap]{proposition}{zerogap} \label{prop:zeroaware}
  Under the assumptions of~\Cref{prop:amgm_gap} with common $\beta>0$,
  a zero-aware replacement for the simple upper bound of
 ~\Cref{cor:simple_amgm} is:
  \begin{equation}
    \label{eq:zerogap}
    \widetilde{\gA}(q,\valpha,\beta,m)\defeq\frac{m}{2}\;
    \Var^{\omega_0}(\valpha){\gA}\bigl(q;\bar \alpha,\beta,m\bigr).
  \end{equation}
\end{restatable}
\noindent
In particular, coordinates with $\alpha_i\equiv 0$ incur zero penalty, while
$\alpha_i=1$ retains full influence through $\Var^{\omega_0}(\valpha)$; when
$\omega_0\!\equiv\!1$ we recover~\Cref{cor:simple_amgm}. More details are provided in~\Cref{appendix:zeroaware} about the derivation of various quantities.

The main takeaway from~\Cref{eq:zerogap} is that the AM-GM gap can be reduced by pushing the active
entries towards a common non-zero value. That is particularly true for the case where
$\alpha_i\in\lrcb{0,1}$ and validates the claim that PRCut~\citep{prcut} bound is tight in the
deterministic setting.

\subsection{Concentration of batch-based estimators} \label{sec:concentration}
Fix $m\in\mathbb{N}$, $q>0$, $\beta>0$. Let $c=\tfrac{q}{\beta}$ and $\valpha \in [0,1]^m$ with
mean $\bar\alpha$ and population variance $\Var(\valpha)$. Form a random minibatch
$S=(i_1,\dots,i_B)$ of size $B$, sampled with replacement, and define the plug-in estimator:
\begin{align}
  \hat{\alpha}_S \;\defeq\; \tfrac{1}{B}\sum_{r=1}^B \alpha_{i_r},
  \qquad
  \hat H(S) \;\defeq\; \gH_\beta\!\bigl(q;\hat{\alpha}_S,m\bigr).
\end{align}
By~\Cref{lem:2F1-lip}, $\gH_\beta(q;\cdot,m)$ is decreasing,
convex and $L$-Lipschitz with:
\begin{equation}
  \label{eq:Lipschitz-H}
  L  \;=\; \frac{m}{q\left(c+1\right)}.
\end{equation}
Differentiating~\Cref{eq:2F1-derivative} again gives:
\begin{align}
  \label{eq:curvature-H}
  \frac{d^2}{dz^2}\,\gH_\beta(q;z,m)
  \;\le\; \underbrace{\frac{1}{q}\cdot \frac{2m(m-1)}{c(c+1)}}_{\!\!\!\defeq K},
\end{align}
and the Taylor bound from~\Cref{eq:curvature-H} yields:
\begin{equation}
  \label{eq:bias-bound}
  0 \;\le\; \Ea{\hat H(S)}-\gH_\beta\!\bigl(q;\bar\alpha,m\bigr)
  \;\le\; \frac{K}{2}\,\frac{\sigma^2}{B}.
\end{equation}

Changing one element of $S$ changes $\hat{\alpha}_S$ by at most $1/B$, hence by
\Cref{eq:Lipschitz-H} the function $S\mapsto \tilde H(S)$ changes by at most $L/n$. We obtain via
McDiarmid's inequality for any $\varepsilon>0$:
\begin{equation}
  \label{eq:mcdiarmid}
  \Pr\!\left(\bigl|\hat H(S)-\Ea{\hat H(S)}\bigr|\!\ge \varepsilon\right)
  \! \le \!2\exp\!\left(\frac{-2B\,\varepsilon^2}{L^2}\right).
\end{equation}

Combining~\Cref{eq:bias-bound,eq:mcdiarmid} with a triangle inequality yields the following
finite-sample guarantee.

\begin{proposition}[Concentration of the minibatch envelope]
  \label{prop:mini-concentration}
  With probability at least $1-\delta$,
  \begin{equation}
    \label{eq:final-conc-bound}
    \bigl|\hat H(S)-\gH_\beta(q;\bar\alpha,m)\bigr|
    \le
    L\,\sqrt{\frac{1}{2B}\log\!\frac{2}{\delta}}
    + \frac{K}{2}\,\frac{\sigma^2}{B},
  \end{equation}
  where $L$ and $K$ defined in~\Cref{eq:Lipschitz-H,eq:curvature-H}.
\end{proposition}

\begin{proof}
  By McDiarmid, with probability $\ge 1-\delta$,
  $|\hat H(S)-\Ea{\hat H(S)}|\le L\sqrt{\tfrac{1}{2B}\log(2/\delta)}$.
  Add and subtract $\gH_\beta(q;\bar\alpha,m)$ and use
  $\Ea{\hat H(S)}-\gH_\beta(q;\bar\alpha,m)\le \tfrac{K}{2}\Var(\hat{\alpha}_S)$
  from~\Cref{eq:bias-bound}, with $\Var(\hat{\alpha}_S)=\sigma^2/B$.
\end{proof}

\subsection{Heterogeneous degrees} \label{sec:hetero-bins}

When $(\beta_i)_i$ vary, directly using a single $\beta$ loses heterogeneity. We partition indices
into $d$ disjoint bins $S_1,\dots,S_d$ based on their $\beta_i$ values. Let $m_j\defeq |S_j|$,
$\bar\alpha_j\defeq m_j^{-1}\sum_{i\in S_j}\alpha_i$, and define the bin interval
$B_j=[b_{j-1},b_j]$ with representatives $\beta_j^\star\in B_j$ specified below.

For $t\in[0,1]$ and $\alpha\in[0,1]$, the map $\beta\mapsto (1-\alpha+\alpha\,t^\beta)$ is
nonincreasing. Hence, for any fixed bin $S_j$ and any choice $\beta_j^\star\le \beta_i$ for all
$i\in S_j$, we have:
\begin{equation}
  \label{eq:beta-monotone}
  \prod_{i\in S_j}\Bigl(1-\alpha_i+\alpha_i\,t^{\beta_i}\Bigr)
  \;\le\;
  \prod_{i\in S_j}\Bigl(1-\alpha_i+\alpha_i\,t^{\beta_j^\star}\Bigr).
\end{equation}
Applying Jensen in $\alpha$ to the RHS (log is concave in $\alpha$ for fixed $t^\beta$)
gives, for each $j$,
\begin{equation}
  \label{eq:bin-amgm}
  \prod_{i\in S_j}\Bigl(1-\alpha_i+\alpha_i\,t^{\beta_j^\star}\Bigr)
  \;\le\;
  \Bigl(1-\bar\alpha_j+\bar\alpha_j\,t^{\beta_j^\star}\Bigr)^{m_j}.
\end{equation}

Recall the envelope $\gH_\beta(q;\bar\alpha,m)$ from~\Cref{thm:hyperbound}. We now control the
heterogeneous case:

\begin{theorem}[Binned Hölder bound]
  \label{thm:holder}
  Let $q>0$ and partition $\{1,\ldots,m\}$ into bins $S_1,\dots,S_d$. Choose
  representatives $\beta_j^\star\in B_j$ satisfying $\beta_j^\star\le \beta_i$ for every
  $i\in S_j$ (e.g., left endpoints). Then
  \begin{equation}
    \label{eq:holder-bound}
    \gI\bigl(q;\valpha,\vbeta\bigr)
    \;\;\le\;\;
    \prod_{j=1}^{d}\Bigl[\;\gH_{\beta_j^\star}\!\bigl(q;\bar\alpha_j,m\bigr)\;\Bigr]^{\frac{m_j}{m}}.
  \end{equation}
\end{theorem}

Hölder inequality is tight iff the functions $\{f_j\}_{j=1}^d$ are pairwise proportional
(\emph{colinear}) in $L^{p_j}$: there exist constants $\kappa_j>0$ and a common shape $\phi$ such
that $f_j(t)=\kappa_j\,\phi(t)$ for almost every $t\in[0,1]$. In our construction,
\[
  f_j(t)\propto t^{\frac{q-1}{p_j}}\Bigl(1-\bar\alpha_j+\bar\alpha_j\,t^{\beta_j^\star}\Bigr)^m.
\]
Hence near-tightness is promoted when, \emph{across bins}, the curves $t\mapsto
  (1-\bar\alpha_j+\bar\alpha_j\,t^{\beta_j^\star})$ have similar shapes, and, \emph{within bins},
replacing $\beta_i$ by $\beta_j^\star$ induces minimal distortion.

\subsection{Optimization objective}
\label{sec:final-objective}
We now put everything together to define the optimization objective of our probabilistic
graph cut framework. For cluster $\ell$, the expected contribution of edge $(i,j)$ is:
\begin{align*}
  \frac{1}{s_i}\mW_{ij}\,\mP_{i\ell}(1-\mP_{j\ell})
  \gI\!\bigl(s_i;\mP_{-\{i\ell\}},\vs_{-\{il\}}\bigr)
\end{align*}

Fix $\ell\in\{1,\dots,k\}$ and partition indices into $d$ bins $S_{\ell 1},\ldots,S_{\ell d}$ by
their exponents $\beta_u\equiv s_u$ (e.g., degree-based); let $m_{\ell j}\defeq |S_{\ell j}|$,
$m_\ell\defeq\sum_j m_{\ell j}$, and
\[
  \bar p_{\ell j}\;\defeq\;\frac{1}{m_{\ell j}}\sum_{u\in S_{\ell j}}\mP_{u\ell},
  \qquad
  w_{\ell j}\;\defeq\;\frac{m_{\ell j}}{m_\ell}.
\]
Choose representatives $\beta_{\ell j}^\star\le s_u$ for all $u\in S_{\ell j}$ (e.g., the bin’s
left endpoint or in–bin minimum) so that the bound direction is preserved (\Cref{sec:hetero-bins}).

For a fixed $q>0$,~\Cref{thm:holder} yields the per–cluster integrand bound. Plugging $q=s_i$ for
each source vertex $i$ gives the \emph{per-vertex} envelope:
\[
  \Phi_\ell(q)\;\defeq\;\prod_{j=1}^{d}
  \Bigl[\;\gH_{\beta_{\ell j}^\star}\!\bigl(q;\bar p_{\ell j},\,m_\ell\bigr)\;\Bigr]^{w_{\ell j}}.
\]
Define the edge–aggregated source weights
\[
  M_{i\ell}(\mP)\;\defeq\;\sum_{j=1}^n \mW_{ij}\,\mP_{i\ell}\,\bigl(1-\mP_{j\ell}\bigr),
\]
so that the total contribution of cluster $\ell$ is:
\begin{equation}
  \label{eq:U-ell}
  U_\ell(\mP)\defeq\sum_{i=1}^n M_{i\ell}(\mP)\,\Phi_\ell\!\bigl(s_i\bigr),
\end{equation}
and $U(\mP)\defeq\sum_{\ell=1}^k U_\ell(\mP)$.
By construction (linearity of expectation and~\Cref{thm:holder}),
$I_{\mathrm{true}}\!\le U$.

Within each bin, replacing $\{\mP_{u\ell}\}_{u\in S_{\ell j}}$ by their mean $\bar p_{\ell j}$
induces an AM--GM gap controlled by~\Cref{prop:amgm_gap,prop:zeroaware}. A conservative, separable
upper bound for cluster $\ell$ is:
\begin{equation}
  \label{eq:Gamma-ell-edge}
  \Gamma_\ell(\mP)\defeq\sum_{i=1}^n M_{i\ell}(\mP)\,
  \Biggl[\;\sum_{j=1}^d w_{\ell j}\, \sA(\beta_{\ell j}^\star, \vp_{\ell j}, m_\ell) \Biggr],
\end{equation}
and $\sA$ is the zero–aware coefficient from~\Cref{prop:zeroaware}.
Summing over clusters,
$\Gamma(\mP)\defeq\sum_{\ell=1}^k \Gamma_\ell(\mP)$ satisfies
\[
  0\;\le\; U(\mP)-I_{\mathrm{true}}(\mP)\;\le\;\Gamma(\mP).
\]

We minimize a penalized majorizer of the expected GraphCut:
\begin{equation}
  \label{eq:final-objective}
  \boxed{
    \min_{\;\mP=\mP(\vz)} \sJ_\rho(\mP)\defeq U(\mP)\;+\;\rho\,\Gamma(\mP),
    \quad \rho\ge 0,\quad}
\end{equation}
where $\vz$ can be the parameterization logits (via Softmax). Since $I_{\mathrm{true}}\le
  U$ and $\Gamma\ge 0$, we retain $I_{\mathrm{true}}\le \sJ_\rho$ for all $\rho\ge0$ while
explicitly shrinking the AM--GM gap.

We detail in ~\Cref{appendix:ForwardBackward} the forward-backward derivation and implementation
for our final objective.

\paragraph{Time complexity.}
With minibatches of size $B$, we first construct the batch adjacency
$\mW_{\text{batch}}\in\R^{B\times B}_+$. For dense similarities this costs $O(B^2)$ time (and
$O(B^2)$ memory); in sparse $k$NN settings we can replace $B^2$ by
$\mathrm{nnz}(\mW_{\text{batch}})$. We then precompute the envelope terms
$\gH_{\beta^\star_j,\ell}$ for every (bin $j$, cluster $\ell$). A straightforward implementation
performs $O(d\,k\,m)$ work, where $d$ is the number of bins, $k$ the number of clusters, and $m$
the polynomial degree in the ${}_2F_1(-m,\cdot\,;\cdot\,;z)$ evaluation. Because these computations
factor across $(j,\ell)$, they are embarrassingly parallel; with $(d\times k)$ workers the
wall-clock reduces to $O(m)$ (see~\Cref{appendix:ForwardBackward}). Thus, each batch step
practically takes $O(\mathrm{nnz}(\mW_{\text{batch}}) \cdot k + m)$.

%% file: related.tex
Let $\rva\in\{0,1\}^n$ be a cluster indicator and $\vp\in[0,1]^n$ with $p_i\defeq\Pr[\rva_i{=}1]$.
Because $p_i^2 \le p_i$ on $[0,1]$, the Laplacian quadratic $\vp^\top\mL\vp$ and the expected cut
$\Ea{\rva^\top\mL\rva}$ coincide \emph{only} when $\vp$ is binary: the degree term satisfies
$\vp^\top\mD\vp = \sum_i d_i p_i^2 \le \sum_i d_i p_i = \Ea{\rva^\top\mD\rva}$, and the adjacency
term $\Ea{\rva^\top\mW\rva} = \vp^\top\mW\vp$ requires the additional assumption
$\Ea{\rva_i\rva_j}=p_ip_j$. Our formulation therefore keeps the expected (linear) degree term
$\sum_i d_i p_i$ rather than the quadratic $\sum_i d_i p_i^2$.

Writing the soft cut for cluster $\ell$ with $\vp = \mP_\ell$:
\[
  \cut_\ell(\vp) = \sum_{ij}\mW_{ij}\,p_i(1{-}p_j)
  = \underbrace{\vp^\top\mL\,\vp}_{\text{convex}}
  + \underbrace{\textstyle\sum_i d_i\,p_i(1{-}p_i)}_{\text{concave}},
\]
a Difference of Convex (DC) decomposition since $\vp^\top\mL\vp$ is convex, while the concave
\emph{fuzziness} term $\sum_i d_i p_i(1{-}p_i)$ vanishes for hard assignments.

Spectral NCut~\citep{standardspectral} relaxes indicators to continuous vectors on the Stiefel
manifold while our approach operates directly on the assignment polytope $\mathcal{U}_{n,K}$, where
the row-stochastic constraint $\sum_\ell \mP_{i\ell}=1$ couples all $K$ columns and precludes the
per-column eigendecomposition.

\paragraph{XOR similarity and cross-entropy relaxation.}
\label{sec:contrast}
The per-edge cut contribution $p_i(1{-}p_j)$ can be read as an \emph{XOR
similarity}: it is maximal when exactly one of $p_i,p_j$ is~1 (a
cross-partition edge) and zero when both agree. Symmetrizing gives the
per-edge cut cost
\[
  \delta_{ij} \;\defeq\; p_i(1{-}p_j)+p_j(1{-}p_i).
\]
Since $1-p\le -\log p$ for $p\in(0,1]$, each term is bounded by its
cross-entropy counterpart, yielding
\[
  \delta_{ij}
  \;\le\; -p_i\log p_j - p_j\log p_i
  \;=\; \mathrm{CE}(p_i\|p_j)+\mathrm{CE}(p_j\|p_i).
\]
Weighting by $\mW_{ij}$ and summing over edges upper-bounds the expected
cut in~\Cref{eq:gcut_prob}. It suffices to consider one direction:
\[
  \mathrm{CE}(p_i\|p_j) = H(p_i) + \kl{p_i}{p_j},
\]
so minimizing the cross-entropy surrogate simultaneously encourages
\emph{neighbor agreement} (small $D_{\mathrm{KL}}$) and \emph{sharp
assignments} (small entropy $H$).

\paragraph{Temperature annealing.}
Parameterizing assignments via a softmax with temperature,
$\mP_{i\ell}=\softmax(z_{i\ell}/\tau)$, provides a smooth
interpolation between uniform ($\tau\to\infty$) and hard
($\tau\to 0$) assignments. As $\tau$ decreases, the probabilities are
pushed toward $\{0,1\}$, which has two effects: (i)~the fuzziness term
$\sum_i d_i p_i(1{-}p_i)$ in the DC decomposition above vanishes, so
$\vp^\top\mL\vp \to \Ea{\rva^\top\mL\rva}$; and (ii)~the gap
$1{-}p \le -\log p$ tightens, so the cross-entropy surrogate converges
to the XOR cut cost. When the cluster index $\ell$ ranges over batch
instances, $\mathrm{CE}(\mP_i\|\mP_j)=-\log\mP_{j,i}$ recovers
InfoNCE; when $\ell$ indexes prototypes, it enforces code consistency.

\paragraph{SimCLR as a probabilistic cut.}
\label{sec:simclr-as-special-case}
SimCLR~\citep{chen20simclr} builds representations via contrastive
learning~\citep{ContrastiveLossLecun}: for each image, two augmented views
are embedded and trained with InfoNCE to attract positive pairs and repel
negatives. This naturally defines a \emph{view graph} with
$\mW_{ab}\defeq\kappa(z_a,z_b)$ where
$\kappa(u,v)=\exp(\langle u,v\rangle/\tau)$. With a single bin ($d{=}1$)
and $K$ equal to the number of latent classes, our envelope $U_\ell$ is
modulated by $\gH(s_i;\bar p_\ell,m_\ell)$, which is decreasing in $\bar
  p_\ell$ (\Cref{lem:2F1-lip}): increasing same-class agreement
monotonically decreases the bound. SimCLR’s alignment and uniformity
objectives reduce cross-edge mass in this view graph, so $\sum_\ell U_\ell$
decreases monotonically under InfoNCE updates. In the instance-discrimination
limit ($K$ equals batch size), minimizers of $\sJ_\rho$ coincide with those
of SimCLR up to the reparameterization $\mP=\softmax(\vz)$.

\paragraph{CLIP as a bipartite cut.}
CLIP~\citep{radford21clip} trains image and text encoders $f_{\theta_x}$, $g_{\theta_t}$ with
symmetric InfoNCE over both directions. Let $z_i=f_{\theta_x}(x_i)$, $u_j=g_{\theta_t}(t_j)$, and
$\kappa(u,v)=\exp(\langle u,v\rangle/\tau)$. We construct a bipartite similarity graph
$\gG=(\gV_x\cup\gV_t,\gE)$ with no intra-modal edges:
\[
  \mW=\begin{pmatrix} 0 & \mW^{xt}\\ (\mW^{xt})^\top & 0 \end{pmatrix},
  \quad \mW_{ij}^{xt}\defeq\kappa(z_i,u_j).
\]
Each text node $t_j$ defines a cluster $\ell{=}j$, and the model learns soft assignments
$\mP_{i\ell}$ of images to text clusters (and symmetrically, texts to image clusters). A matched
image--text pair $(x_i,t_j)$ with $y_i{=}j$ should have high $\mP_{ij}$; minimizing the bipartite
cut reduces the cross-edge mass $\sum_{(i,j):y_i\ne j}\mW_{ij}^{xt}$, driving mismatched
similarities down.

Because the two modalities live in different representation spaces, we use $d=2$ H\"older bins
with representatives $\beta_x^\star$, $\beta_t^\star$ (\Cref{sec:hetero-bins}). The per-cluster
envelope then factorizes as a product of the image-side and text-side envelopes. Taking the $\log$
converts this product into a sum over the two directions, recovering CLIP’s symmetric
image$\to$text and text$\to$image InfoNCE structure. In the paired-supervision limit (one text per
class, one image per class), the minimizers coincide with those of CLIP up to the
reparameterization $\mP=\softmax(\vz)$.

%% file: results.tex
\input{./results_fig.tex}
\section{Experiments}
\input{./results_tables.tex}
We verify our bounds on a synthetic dataset of three intertwined helices with Gaussian noise and
unbalanced clusters ($|C_1|{=}200$, $|C_2|{=}|C_3|{=}400$;~\Cref{fig:helices}). A 50-NN RBF graph
captures the manifold topology. We simulate soft assignments $\mP_{i\ell}=\softmax(z_{i\ell}/\tau)$
at varying temperatures and estimate the expected cut via Monte Carlo (MC) sampling.
\Cref{fig:binning_compare} shows that log-adaptive binning consistently produces a tighter H-NCut
bound than equal-frequency binning. Additional MC simulations for RCut and NCut, per-cluster
breakdowns, and the effect of the number of bins are reported in~\Cref{app:binning_sim}.

\paragraph{Similarity Quality}\label{para:sim-quality}
The performance of graph-based clustering depends critically on the quality of the constructed
similarity graph. We introduce a normalized metric to quantify this:
\begin{definition}[Graph Quality]
  Given a similarity graph $\mW$ and ground-truth labels $y$, let $\mT = \mD^{-1}\mW$
  be the random-walk transition matrix. The \emph{graph quality} is:
  \begin{equation}
    Q = \frac{q - q_{\text{chance}}}{1 - q_{\text{chance}}}, \quad
    q = \frac{1}{n}\sum_{i=1}^n \sum_{j=1}^n \mT_{ij} \cdot \one[y_i = y_j],
  \end{equation}
  where $n_k = |\{i : y_i = k\}|$ is the size of class $k$ and
  $q_{\text{chance}} = \sum_k (n_k/n)^2$ is the probability of a same-class
  transition under a random labeling.
\end{definition}

$Q=0$ means the graph carries no class information beyond chance; $Q=1$ means a single random walk step always
lands in the same class. This metric accounts for the number of classes: $q = 0.1$ is excellent for $K=100$ (since
$q_{\text{chance}} \approx 0.01$) but poor for $K=2$ (where $q_{\text{chance}} \approx 0.5$).

\begin{figure}[t]
  \centering
  \includegraphics[width=0.8\columnwidth]{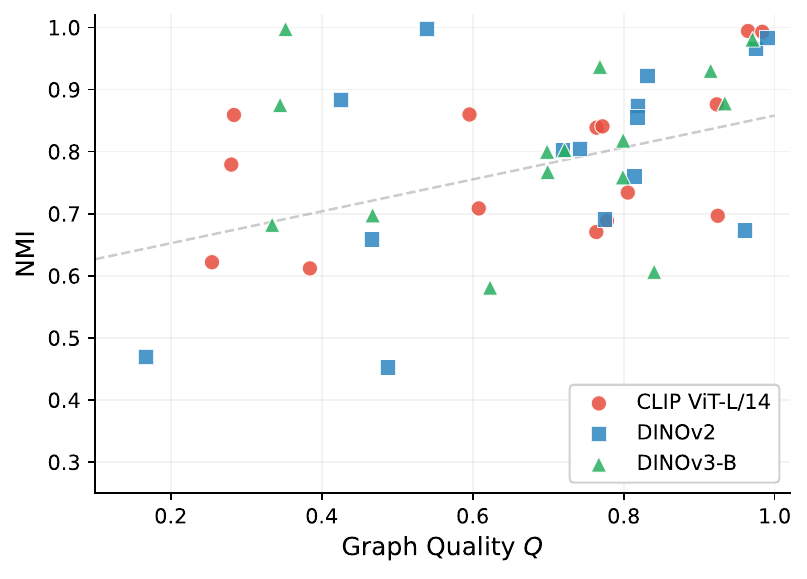}
  \caption{Graph quality $Q$ vs.\ best NMI achieved by H-Cut across 15 datasets and three embedding models.
    The strong linear correlation confirms that $Q$ is a reliable predictor of downstream clustering performance.}
  \label{fig:nmi_vs_quality}
\end{figure}
We observe a strong correlation between $Q$ and downstream clustering accuracy across
all datasets and embedding models (\Cref{tab:dinov2,tab:dinov3b,tab:clipvitL14}),
confirming that \emph{the graph is the bottleneck}: when $Q > 0.8$, H-Cut consistently
achieves high accuracy; when $Q < 0.4$, no graph-cut method can recover the class structure.

To apply the Binned H\"older Bound (Theorem 2) effectively, we must partition the vertices $\{1,
  \dots, n\}$ into $d$ disjoint sets $S_1, \dots, S_d$ such that the representative exponent
$\beta_j^\star = \min_{i \in S_j} \beta_i$ provides a tight lower bound for all $\beta_i$ in that
bin. We consider two strategies \textbf{Equal-Frequency Binning}, \textbf{Log-Adaptive (K-Means)
  Binning} (see~\Cref{app:binning}).

\subsection{Evaluation and Baselines}
We evaluate performance on standard benchmarks: CIFAR-10/100~\citep{cifar}, STL-10~\citep{stl10},
EuroSAT~\citep{eurosat}, MNIST~\citep{mnist}, FashionMNIST~\citep{fashionmnist}, using both
clustering metrics, Accuracy (ACC) and
Normalized Mutual Information (NMI); and geometric graph metrics (Ratio Cut and Normalized Cut). We
compare against: Logistic Regression (serving as a topological upper bound). K-Means, Spectral
Clustering (Normalized Laplacian followed by K-Means), PRCut~\citep{prcut}, and our hypergeometric
objective for $\rcut$ and $\ncut$.

Experimental details, hyperparameters, and additional results are provided
in~\Cref{app:experimental}. We benchmark performance across 15 datasets using three foundation
model embeddings: CLIP ViT-L/14~\citep{radford21clip}, DINOv2~\citep{dinov2}, and
DINOv3-B~\citep{simeoni25dinov3} (\Cref{tab:dinov2,tab:dinov3b}; CLIP
in~\Cref{tab:clipvitL14}). We compare three probabilistic cut objectives: PRCut~\citep{prcut}, our
hypergeometric RatioCut (H-RCut,~\Cref{thm:hyperbound}), and hypergeometric NCut with
H\"older binning (H-NCut, Theorem~2). All methods use the same linear model $\mP =
  \softmax(\mW_\theta \vx / \tau)$ with edge-pair sampling and gradient mixing
(\Cref{app:gradient_mixing}). We denote PRCut trained with our gradient mixing strategy as
PRCut$^*$ since the original PRCut needs slower training schedule to avoid cluster collapse.

\paragraph{Key observations:} Non-parametric SC generally achieves lower RCut/NCut values, but
the parametric methods (PRCut$^*$, H-RCut, H-NCut) consistently achieve higher ACC/NMI: the linear
model acts as an implicit regularizer, preventing overfitting to noisy edges.
Among the parametric methods, no single objective dominates: PRCut$^*$ excels on well-separated
embeddings (Flowers, Pets), while H-NCut stands out on larger $K$ where volume normalization matters
(CIFAR-100, STL-10). The choice of embedding model is at least as important as the algorithm:
DINOv2 achieves 98.6\% on CIFAR-10 while CLIP reaches only 89.6\%. (4) The graph quality metric $Q$
(\Cref{para:sim-quality}) strongly predicts downstream accuracy, confirming that the similarity
graph construction is the primary bottleneck.

%% file: results_fig.tex
\begin{figure*}[!t]
  \centering
  \begin{subfigure}[b]{0.3\textwidth}
    \centering
    \includegraphics[width=0.7\textwidth]{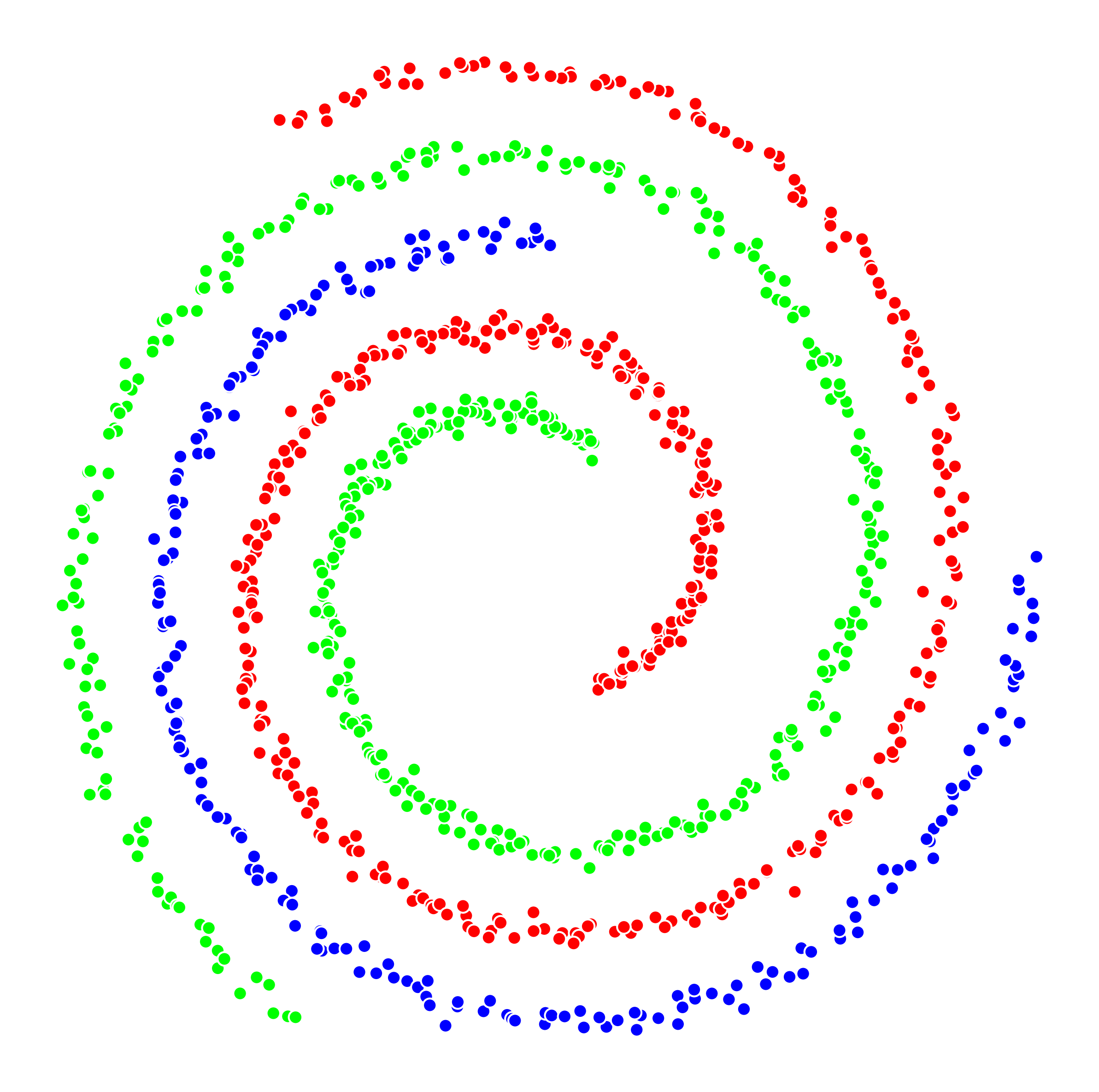}
    \caption{Helix clusters of sizes $(200,400,400)$}
    \label{fig:helices_true}
  \end{subfigure}
  \hfill
  \begin{subfigure}[b]{0.3\textwidth}
    \centering
    \begin{overpic}[width=0.7\textwidth]{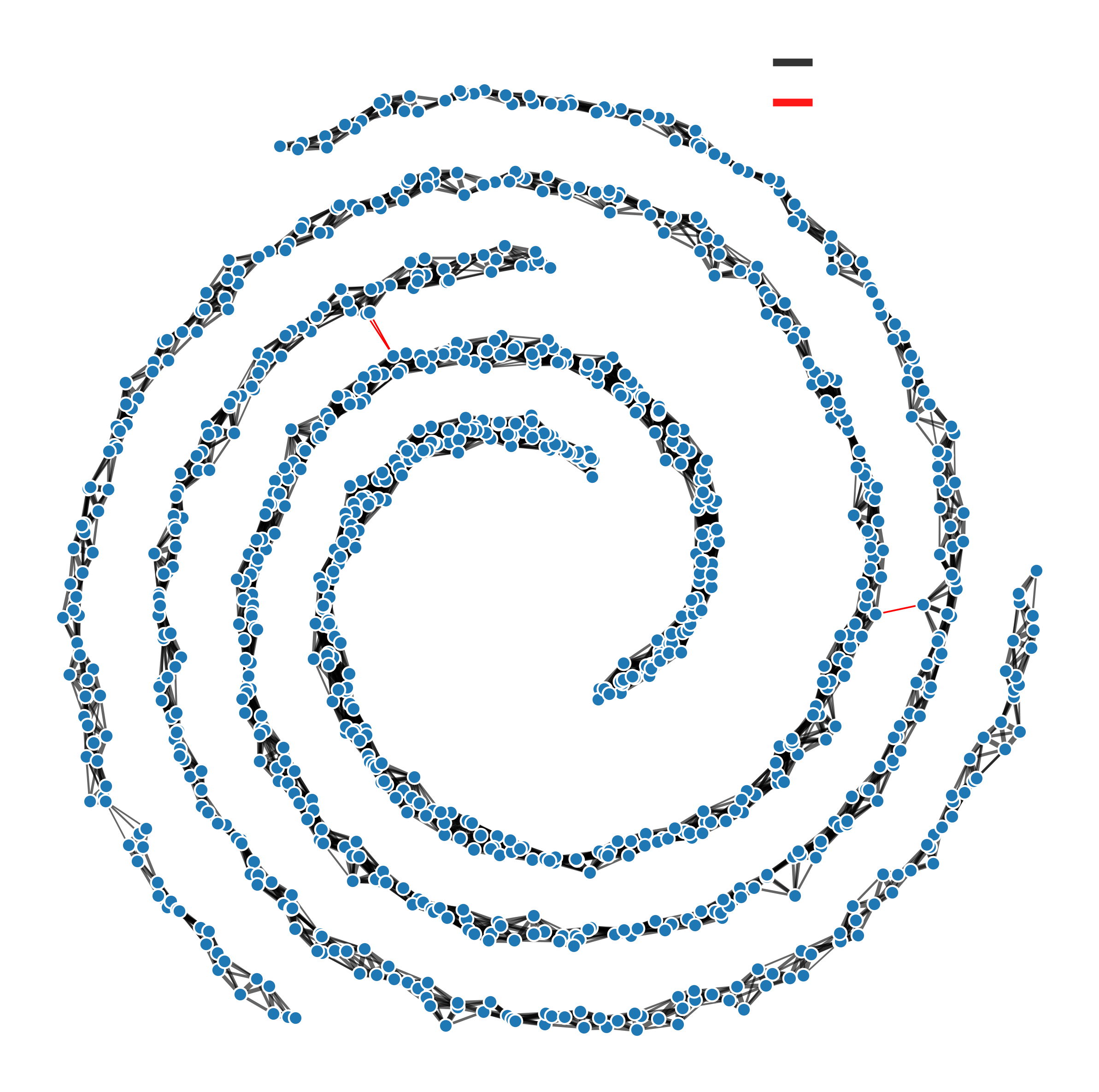}
      \put(74,93){{\tiny intra-class edge}}
      \put(74,89){{\tiny inter-class edge}}
    \end{overpic}
    \caption{constructed graph adjacency}
    \label{fig:helices_graph}
  \end{subfigure}
  \hfill
  \begin{subfigure}[b]{0.3\textwidth}
    \centering
    \includegraphics[width=0.7\textwidth]{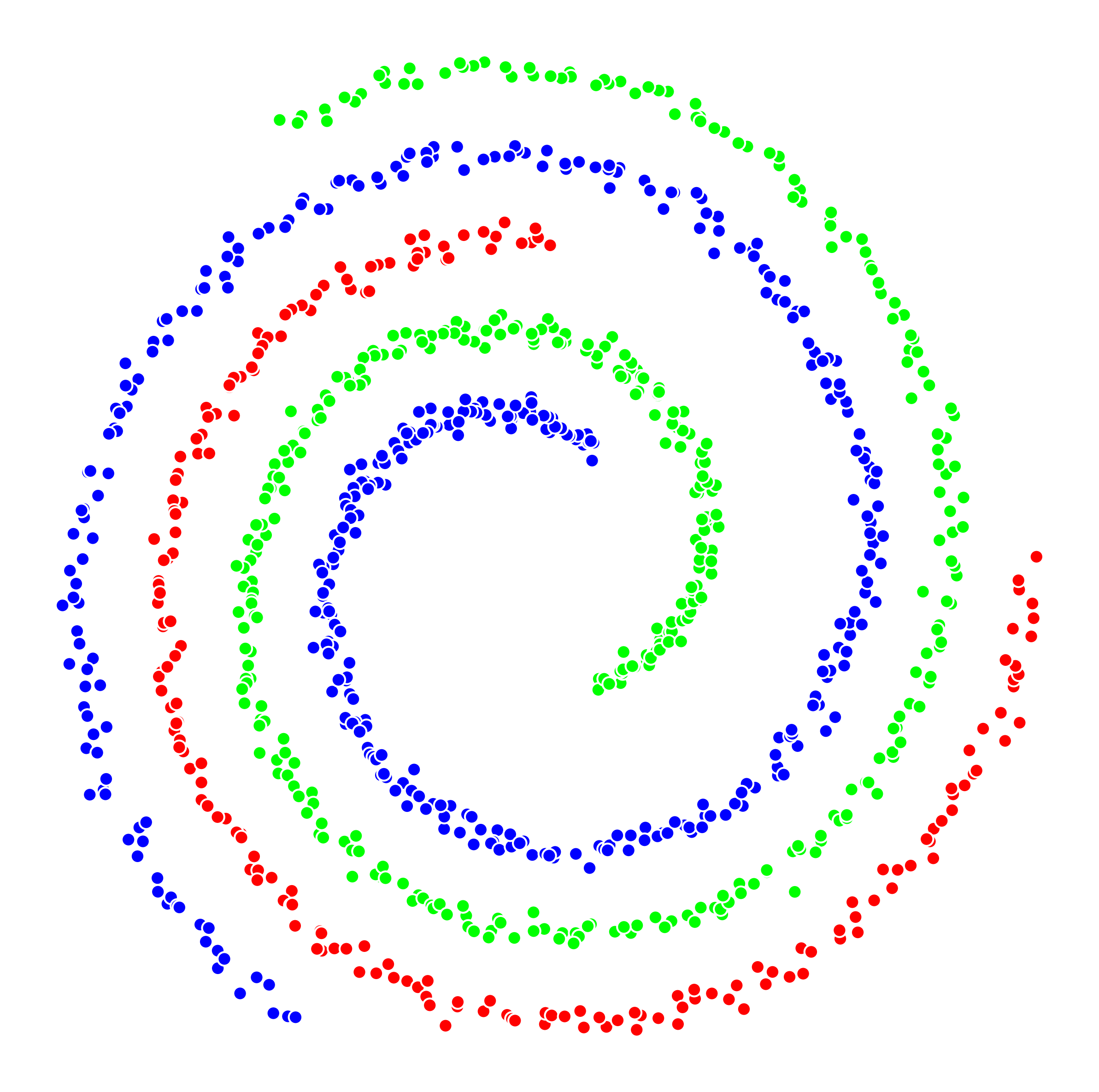}
    \caption{Spectral Clustering solution}
    \label{fig:helices_sc}
  \end{subfigure}

  \vspace{0.3cm}

  \begin{subfigure}[b]{0.32\textwidth}
    \centering
    \includegraphics[width=\textwidth]{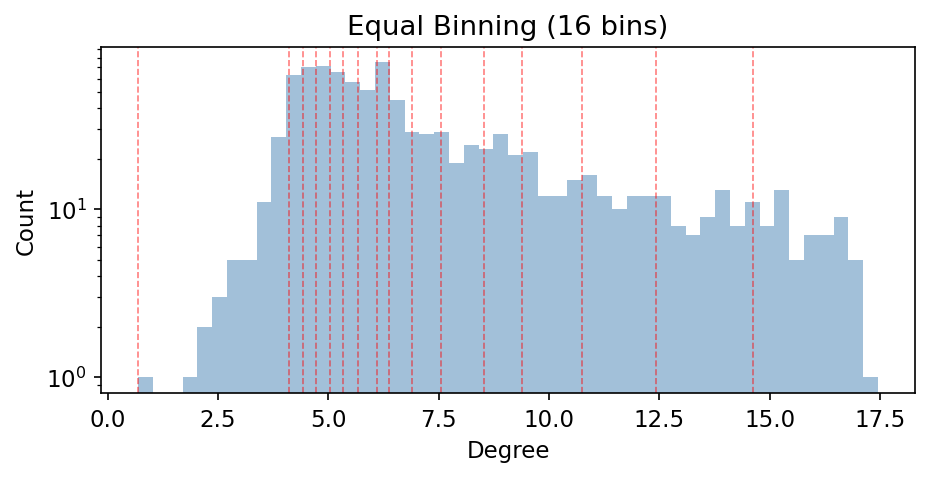}
    \caption{Equal-frequency binning ($d{=}16$)}
    \label{fig:equal_binning}
  \end{subfigure}
  \hfill
  \begin{subfigure}[b]{0.32\textwidth}
    \centering
    \includegraphics[width=\textwidth]{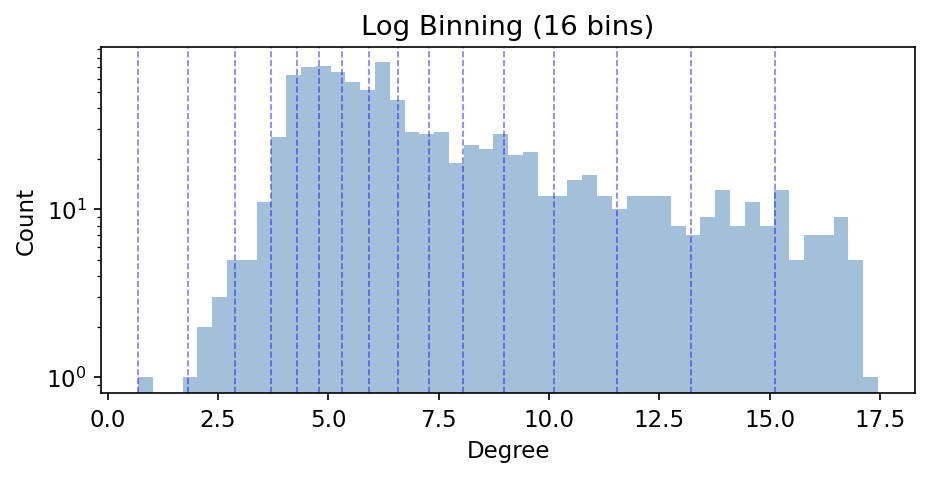}
    \caption{Log-adaptive binning ($d{=}16$)}
    \label{fig:log_binning}
  \end{subfigure}
  \hfill
  \begin{subfigure}[b]{0.32\textwidth}
    \centering
    \includegraphics[width=\textwidth]{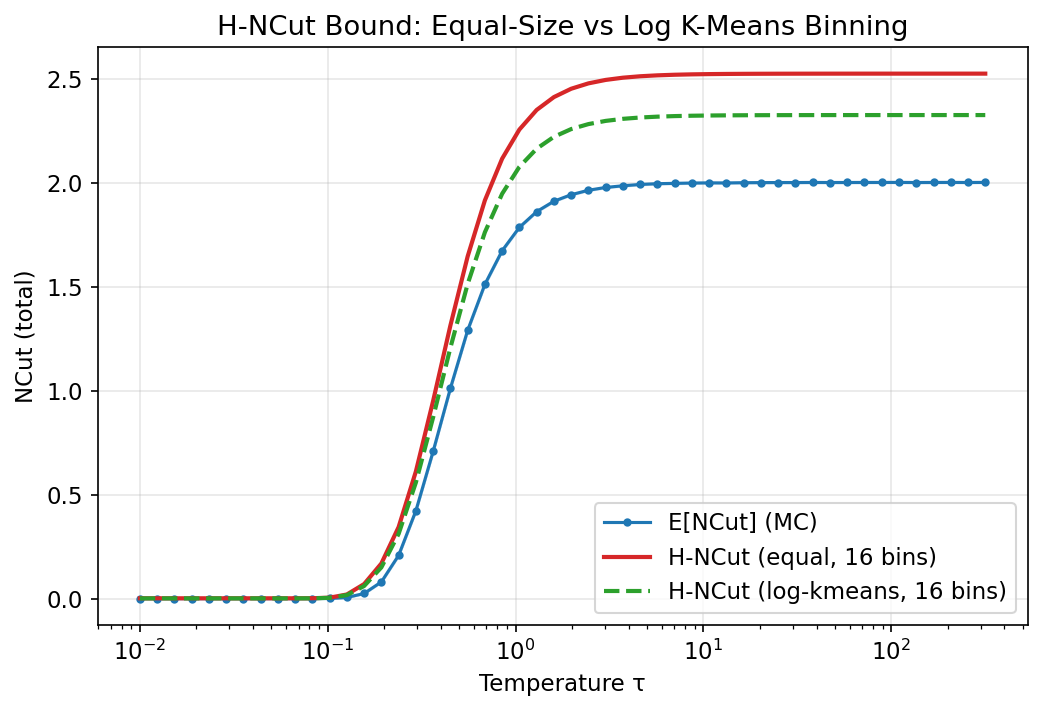}
    \caption{H-NCut bound: equal vs.\ log-adaptive}
    \label{fig:binning_compare}
  \end{subfigure}

  \caption{\textbf{Synthetic helix simulation.} \emph{Top:} three intertwined helices with
    unbalanced clusters and the constructed KNN graph. \emph{Bottom:} degree distribution with
    bin boundaries for equal-frequency (d) and log-adaptive (e) binning. Log-adaptive binning
    concentrates boundaries where the degree distribution has mass, producing a tighter H-NCut
    bound across all temperatures (f). See~\Cref{app:binning_sim} for per-cluster MC simulations.}
  \label{fig:helices}
\end{figure*}

%% file: results_tables.tex
\newcommand{\mcf}[1]{ \multicolumn{4}{c|}{#1}}
\newcommand{\mcfl}[1]{ \multicolumn{4}{c}{#1}}
\newcommand{\mcfsc}{\mcf{Spectral Clustering}}
\newcommand{\mcfpr}{\mcf{PRCut$^*$}}
\newcommand{\mcfhr}{\mcf{H-RCut}}
\newcommand{\mcfhn}{\mcfl{H-NCut}}

\begin{table*}[t]
  \centering
  \caption{\textbf{DINOv2 embeddings~\citep{dinov2}.} Non-parametric Spectral Clustering (SC) vs.\ parametric probabilistic cut objectives on a 50-NN RBF graph. SC directly optimizes the graph Laplacian and achieves the lowest RCut/NCut, but H-Cut methods generalize better in ACC/NMI by leveraging the linear model as an implicit regularizer.}
  \label{tab:dinov2}
  \resizebox{\textwidth}{!}{
    \begin{tabular}{l|r|c|cccc|cccc|cccc|cccc}
      \toprule
                                               &     &      & \mcfsc        & \mcfpr        & \mcfhr            & \mcfhn                                                                                                                                                                                                                                    \\
      \textbf{Dataset}                         & $K$ & $Q$  & ACC\%         & NMI\%         & RCut $\downarrow$ & NCut $\downarrow$ & ACC\%         & NMI\%         & RCut $\downarrow$ & NCut $\downarrow$ & ACC\%         & NMI\%         & RCut $\downarrow$ & NCut $\downarrow$ & ACC\%         & NMI\%         & RCut $\downarrow$ & NCut $\downarrow$ \\
      \midrule
      CIFAR-10~\citep{cifar}             & 10  & 0.98 & 78.5          & 90.2          & \textbf{5.7}      & \textbf{0.12}     & 93.2          & 91.4          & 13.1              & 0.29              & 98.5          & 96.5          & 8.9               & 0.20              & \textbf{98.6} & \textbf{96.6} & 8.8               & 0.20              \\
      CIFAR-100~\citep{cifar}            & 100 & 0.82 & 67.4          & 81.6          & \textbf{188}      & \textbf{4.3}      & 81.7          & 86.3          & 601               & 12.6              & 80.6          & 86.0          & 562               & 11.7              & \textbf{83.8} & \textbf{87.3} & 543               & 11.7              \\
      STL-10~\citep{stl10}               & 10  & 0.96 & 51.2          & 64.0          & \textbf{3.9}      & \textbf{0.09}     & 62.4          & 63.5          & 36.0              & 0.88              & \textbf{68.1} & \textbf{67.3} & 33.9              & 0.83              & 68.0          & \textbf{67.3} & 32.3              & 0.80              \\
      EuroSAT~\citep{eurosat}            & 10  & 0.82 & 80.9          & 78.0          & 54                & 1.2               & \textbf{93.2} & \textbf{85.5} & 52.6              & \textbf{1.1}      & 92.5          & 84.5          & \textbf{52.1}     & \textbf{1.1}      & 92.5          & 84.5          & \textbf{52.1}     & \textbf{1.1}      \\
      Imagenette~\citep{imagenette}      & 10  & 0.99 & \textbf{99.8} & \textbf{99.3} & \textbf{2.4}      & \textbf{0.05}     & 99.2          & 98.1          & 3.5               & 0.08              & 99.2          & 98.3          & 3.3               & 0.07              & 99.2          & 98.3          & 3.2               & 0.07              \\
      Fashion-MNIST~\citep{fashionmnist} & 10  & 0.81 & 71.2          & 75.4          & \textbf{26}       & \textbf{0.58}     & \textbf{77.8} & 75.8          & 38.4              & 0.83              & 77.3          & \textbf{76.0} & 38.0              & 0.83              & 77.1          & 75.9          & 38.3              & 0.83              \\
      MNIST~\citep{mnist}                & 10  & 0.78 & 62.1          & 63.0          & \textbf{60}       & \textbf{1.3}      & 72.6          & 64.4          & 80.9              & 1.8               & \textbf{77.0} & \textbf{69.1} & 73.3              & 1.6               & 72.8          & 65.8          & 77.2              & 1.7               \\
      Pets~\citep{pets}                  & 37  & 0.83 & 87.0          & 91.6          & \textbf{140}      & \textbf{3.6}      & \textbf{90.4} & \textbf{92.2} & 176               & 4.4               & 89.9          & 91.9          & 176               & 4.4               & 90.0          & 91.8          & 172               & 4.3               \\
      Flowers-102~\citep{flowers}        & 102 & 0.54 & 95.7          & 97.8          & \textbf{1985}     & \textbf{49.5}     & \textbf{99.7} & \textbf{99.7} & 2159              & 49.9              & 93.0          & 97.3          & 2386              & 52.3              & 94.3          & 97.2          & 2256              & 50.9              \\
      Food-101~\citep{food101}           & 101 & 0.74 & 71.8          & 79.7          & \textbf{367}      & \textbf{8.0}      & 75.8          & 79.9          & 558               & 12.3              & 76.3          & \textbf{80.4} & 566               & 12.5              & \textbf{76.6} & \textbf{80.4} & 584               & 12.9              \\
      RESISC-45~\citep{resisc45}         & 45  & 0.72 & 68.0          & 74.9          & \textbf{314}      & \textbf{7.3}      & \textbf{79.2} & \textbf{80.2} & 437               & 9.8               & 74.5          & 78.3          & 429               & 9.6               & 78.4          & 79.8          & 419               & 9.4               \\
      DTD~\citep{dtd}                    & 47  & 0.47 & 54.9          & 64.1          & \textbf{575}      & \textbf{15.0}     & \textbf{57.9} & 65.7          & 760               & 18.3              & 55.1          & 64.3          & 843               & 18.1              & 57.8          & \textbf{65.8} & 776               & 17.6              \\
      GTSRB~\citep{gtsrb}                & 43  & 0.49 & 35.2          & 43.7          & \textbf{526}      & \textbf{12.9}     & 36.2          & 44.2          & 760               & 17.3              & \textbf{37.9} & \textbf{45.2} & 745               & 17.1              & 36.7          & 44.1          & 746               & 17.2              \\
      CUB-200~\citep{cub}                & 200 & 0.43 & 68.1          & 86.1          & 4029              & \textbf{107}      & \textbf{70.5} & \textbf{88.3} & \textbf{3784}     & 111               & 58.0          & 83.5          & 5244              & 118               & 58.7          & 83.8          & 5296              & 118               \\
      FGVC-Aircraft~\citep{aircraft}     & 100 & 0.17 & \textbf{21.9} & 46.6          & 1494              & \textbf{38.0}     & 21.4          & \textbf{46.9} & \textbf{1453}     & 42.3              & 20.1          & 45.6          & 1665              & 41.4              & 20.1          & 45.7          & 1639              & 41.3              \\
      \bottomrule
    \end{tabular}}
\end{table*}

\begin{table*}[t]
  \centering
  \caption{\textbf{DINOv3-B embeddings~\citep{simeoni25dinov3}.} DINOv3 uses register tokens and SigLIP-style training, producing higher-quality graphs on coarse-grained datasets (STL-10, GTSRB) but lower $Q$ on fine-grained Flowers and CUB.}
  \label{tab:dinov3b}
  \resizebox{\textwidth}{!}{
    \begin{tabular}{l|r|c|cccc|cccc|cccc|cccc}
      \toprule
                                               &     &      & \mcfsc        & \mcfpr        & \mcfhr            & \mcfhn                                                                                                                                                                                                                                    \\
      \textbf{Dataset}                         & $K$ & $Q$  & ACC\%         & NMI\%         & RCut $\downarrow$ & NCut $\downarrow$ & ACC\%         & NMI\%         & RCut $\downarrow$ & NCut $\downarrow$ & ACC\%         & NMI\%         & RCut $\downarrow$ & NCut $\downarrow$ & ACC\%         & NMI\%         & RCut $\downarrow$ & NCut $\downarrow$ \\
      \midrule
      CIFAR-10~\citep{cifar}             & 10  & 0.93 & 87.5          & \textbf{90.8} & \textbf{14}       & \textbf{0.33}     & 86.6          & 83.7          & 34.9              & 0.83              & \textbf{91.2} & 87.8          & 31.5              & 0.75              & 85.6          & 83.3          & 46.1              & 1.1               \\
      CIFAR-100~\citep{cifar}            & 100 & 0.70 & 63.1          & 76.2          & \textbf{638}      & \textbf{15.7}     & \textbf{73.2} & \textbf{80.0} & 1216              & 26.8              & 69.6          & 77.6          & 1306              & 28.9              & 70.7          & 78.9          & 1324              & 28.8              \\
      STL-10~\citep{stl10}               & 10  & 0.92 & 88.6          & \textbf{93.4} & \textbf{24}       & \textbf{0.62}     & 87.2          & 86.4          & 44.0              & 1.1               & 86.5          & 88.2          & 40.6              & 1.0               & \textbf{95.1} & 93.0          & 31.9              & 0.83              \\
      EuroSAT~\citep{eurosat}            & 10  & 0.80 & 82.7          & 79.3          & \textbf{53}       & \textbf{1.2}      & 89.1          & 81.1          & 68.6              & 1.6               & \textbf{91.0} & \textbf{81.8} & 65.8              & 1.5               & 86.2          & 79.4          & 69.8              & 1.6               \\
      Imagenette~\citep{imagenette}      & 10  & 0.97 & \textbf{99.4} & \textbf{98.3} & \textbf{8.8}      & \textbf{0.20}     & 99.3          & 98.1          & 9.4               & 0.22              & 99.3          & 98.1          & 9.6               & 0.22              & 99.3          & 98.1          & 9.6               & 0.22              \\
      Fashion-MNIST~\citep{fashionmnist} & 10  & 0.80 & 70.3          & 74.1          & \textbf{22}       & \textbf{0.51}     & 77.5          & 74.1          & 43.3              & 0.96              & 78.5          & \textbf{75.9} & 39.3              & 0.87              & \textbf{78.6} & \textbf{75.9} & 39.2              & 0.87              \\
      MNIST~\citep{mnist}                & 10  & 0.84 & \textbf{76.9} & \textbf{75.2} & \textbf{40}       & \textbf{0.92}     & 64.1          & 58.5          & 77.2              & 1.8               & 66.5          & 60.6          & 74.1              & 1.7               & 66.5          & 59.3          & 74.4              & 1.7               \\
      Pets~\citep{pets}                  & 37  & 0.77 & 89.8          & 91.2          & \textbf{256}      & \textbf{6.7}      & \textbf{93.7} & \textbf{93.7} & 279               & 7.2               & 93.3          & 93.4          & 276               & 7.1               & 93.2          & 93.3          & 276               & 7.2               \\
      Flowers-102~\citep{flowers}        & 102 & 0.35 & 94.1          & 97.3          & \textbf{2392}     & \textbf{66.8}     & \textbf{99.7} & \textbf{99.7} & 2451              & 66.5              & 86.5          & 92.3          & 2633              & 69.3              & 86.5          & 93.8          & 2638              & 69.0              \\
      Food-101~\citep{food101}           & 101 & 0.72 & 73.5          & \textbf{80.3} & \textbf{592}      & \textbf{13.3}     & \textbf{76.9} & \textbf{80.3} & 876               & 20.4              & 75.9          & 79.8          & 857               & 19.6              & 74.5          & 79.0          & 908               & 20.8              \\
      RESISC-45~\citep{resisc45}         & 45  & 0.70 & 66.7          & 74.6          & \textbf{363}      & \textbf{8.6}      & \textbf{75.1} & \textbf{76.7} & 497               & 11.6              & 70.7          & 74.6          & 529               & 12.3              & 70.6          & 74.4          & 527               & 12.2              \\
      DTD~\citep{dtd}                    & 47  & 0.47 & 61.0          & 68.3          & \textbf{692}      & \textbf{18.0}     & \textbf{63.3} & \textbf{69.7} & 832               & 20.5              & 60.0          & 67.5          & 868               & 20.9              & 59.7          & 67.6          & 870               & 21.0              \\
      GTSRB~\citep{gtsrb}                & 43  & 0.62 & \textbf{48.0} & \textbf{60.5} & \textbf{364}      & \textbf{9.2}      & 46.1          & 55.6          & 619               & 15.1              & 45.2          & 55.9          & 649               & 15.7              & 46.1          & 58.1          & 627               & 15.1              \\
      CUB-200~\citep{cub}                & 200 & 0.34 & 73.7          & \textbf{88.2} & \textbf{4761}     & \textbf{125}      & \textbf{74.2} & 87.5          & 5061              & 128               & 62.3          & 83.2          & 5431              & 135               & 65.0          & 83.6          & 5298              & 133               \\
      FGVC-Aircraft~\citep{aircraft}     & 100 & 0.33 & \textbf{42.6} & \textbf{69.5} & \textbf{1722}     & \textbf{43.4}     & 39.2          & 68.2          & 1908              & 48.0              & 41.0          & 67.6          & 1968              & 48.8              & 39.6          & 67.5          & 1945              & 48.0              \\
      \bottomrule
    \end{tabular}}
\end{table*}

%% file: appendix.tex
\section{Proofs}
\subsection{Generalized Poisson-Binomial}
\begin{lemma}[PGF of a weighted Bernoulli sum]
  Let $r_i \sim \mathrm{Bernoulli}(\alpha_i)$ be independent and define
  $X \defeq \sum_{i=1}^m \beta_i r_i$ with $\beta_i \in \mathbb{Z}_{\ge 0}$.
  Then the probability generating function $G_X(t)\defeq \mathbb{E}[t^X]$ (for $|t|\le 1$) is
  \[
    G_X(t) \;=\; \prod_{i=1}^m \bigl( (1-\alpha_i) + \alpha_i\, t^{\beta_i} \bigr).
  \]
\end{lemma}

\begin{proof}
  Since $X=\sum_i \beta_i r_i$ and $r_i\in\{0,1\}$,
  \(
  t^X = \prod_{i=1}^m t^{\beta_i r_i}.
  \)
  By independence,
  \[
    G_X(t) \;=\; \mathbb{E}\!\left[\prod_{i=1}^m t^{\beta_i r_i}\right]
    = \prod_{i=1}^m \mathbb{E}\!\left[t^{\beta_i r_i}\right].
  \]
  For each $i$, $\mathbb{E}[t^{\beta_i r_i}] = (1-\alpha_i) t^{0} + \alpha_i t^{\beta_i} =
    (1-\alpha_i) + \alpha_i t^{\beta_i}$. Multiplying the factors yields the claim.
\end{proof}

\subsection{Integral Representation: Proof of Lemma~\ref{lemma:integralexp}}
\label[appendix]{appendix:integralexp}
\integralexp*

\begin{proof}
  The proof uses the integral representation of the reciprocal.
  For any $X>0$, we have $\frac{1}{X}=\int_{0}^{1} t^{X-1}\,dt$.
  Applying this with $X=q+\rx$ (which is a.s.\ positive since $q>0$ and $\rx\ge 0$),
  \[
    \frac{1}{q+\rx}=\int_{0}^{1} t^{\,q+\rx-1}\,dt.
  \]

  Taking expectations and using Tonelli’s theorem (the integrand $t^{\,q+\rx-1}$ is nonnegative on
  $[0,1]$),
  \begin{align*}
    \mathbb{E}\!\left[\frac{1}{q+\rx}\right]
     & = \mathbb{E}\!\left[\int_{0}^{1} t^{\,q+\rx-1}\,dt\right]
    = \int_{0}^{1} \mathbb{E}\!\left[t^{\,q+\rx-1}\right] dt         \\
     & = \int_{0}^{1} \mathbb{E}\!\left[t^{\,q-1}\,t^{\rx}\right] dt
    = \int_{0}^{1} t^{\,q-1}\,\mathbb{E}\!\left[t^{\rx}\right] dt.
  \end{align*}

  Here $\mathbb{E}[t^{\rx}]$ is the probability generating function (PGF) of $\rx$, denoted
  $G_{\rx}(t)$. Substituting the PGF from~\Cref{eq:gpb_pgf} gives
  \[
    \mathbb{E}\!\left[\frac{1}{q+\rx}\right]
    = \int_{0}^{1} t^{\,q-1} G_{\rx}(t)\,dt
    = \int_{0}^{1} t^{\,q-1}\!\left[\prod_{i=1}^m \bigl(1-\alpha_i+\alpha_i t^{\beta_i}\bigr)\right] dt.
  \]
\end{proof}

\subsection{Hypergeometric Bound: Proof of Theorem~\ref{thm:hyperbound}}
\label[appendix]{appendix:hyperbound}
\hyperbound*

\begin{proof}
  Assume $\beta_i\equiv \beta>0$ and $q>0$. Recall the definition of $\gI(q,\valpha,\beta)$:
  \[
    \gI(q,\valpha,\beta) \defeq \int_{0}^{1}\!\Biggl[\prod_{i=1}^m \bigl(1-\alpha_i+\alpha_i t^{\beta}\bigr)\Biggr] t^{q-1}\,dt.
  \]
  For fixed $t\in[0,1]$, the map $\alpha \mapsto \log\!\bigl(1-\alpha+\alpha t^{\beta}\bigr)$ is
  concave (log of a positive affine function), hence by Jensen:
  \[
    \sum_{i=1}^m \log\!\bigl(1-\alpha_i+\alpha_i t^{\beta}\bigr)
    \;\le\; m \log\!\bigl(1-\bar\alpha+\bar\alpha\, t^{\beta}\bigr),
    \quad \bar\alpha \defeq \tfrac{1}{m}\sum_{i=1}^m \alpha_i.
  \]
  Exponentiating and integrating gives:
  \[
    I \;\le\; \int_{0}^{1} \bigl(1-\bar\alpha+\bar\alpha\, t^{\beta}\bigr)^m t^{q-1}\,dt \;=\; B.
  \]
  Evaluate $B$ via $u=t^{\beta}$ (so $dt=\tfrac{1}{\beta}u^{\tfrac{1}{\beta}-1}du$):
  \[
    B=\frac{1}{\beta}\int_{0}^{1} (1-\bar\alpha+\bar\alpha u)^m\,u^{\tfrac{q}{\beta}-1}\,du
    = \frac{1}{\beta}\int_{0}^{1} (1-\bar\alpha v)^m (1-v)^{\tfrac{q}{\beta}-1}\,dv,
  \]
  with $v=1-u$. By Euler’s integral for ${}_2F_1$ with
  $(a,b,c,z)=(-m,1,\tfrac{q}{\beta}+1,\bar\alpha)$ (valid since $c>b>0$),
  \[
    B=\frac{1}{\beta}\cdot\frac{\Gamma(1)\Gamma(\tfrac{q}{\beta})}{\Gamma(\tfrac{q}{\beta}+1)}\,
    {}_2F_1\!\Bigl(-m,1;\tfrac{q}{\beta}+1;\bar\alpha\Bigr)
    =\frac{1}{q}\,{}_2F_1\!\Bigl(-m,1;\tfrac{q}{\beta}+1;\bar\alpha\Bigr).
  \]
  Therefore:
  \[
    \boxed{\ I \;\le\; \frac{1}{q}\,{}_2F_1\!\Bigl(-m,1;\tfrac{q}{\beta}+1;\bar\alpha\Bigr)\ }.
  \]
\end{proof}

\subsection{AM-GM Gap: Proof of Proposition~\ref{prop:amgm_gap}}
\label[appendix]{appendix:amgm_gap}
\amgmgap*

\begin{proof}
  Fix $\tau \in [0,1]$ and set $c \defeq 1-\tau$. Define $f_\tau(\alpha)\defeq
    \log(1-\alpha+\alpha\tau)=\log(1-c\,\alpha)$, a concave function on $[0,1]$ with:
  \[
    f_\tau''(\alpha)= -\,\frac{c^2}{(1-c\,\alpha)^2}\ \in\ \Bigl[-\frac{c^2}{\tau^2},\, -\,c^2\Bigr].
  \]
  Thus $-f_\tau$ is $\theta_\tau$-smooth with $\theta_\tau=c^2/\tau^2$ and $\gamma_\tau$-strongly
  convex with $\gamma_\tau=c^2$ on $[0,1]$. By the standard Jensen two-sided bound for
  twice-differentiable concave functions:
  \begin{equation}
    \label{eq:jensen-two-sided}
    \frac{\gamma_\tau}{2}\,\Var(\valpha)
    \;\le\; f_\tau(\bar \alpha)-\frac{1}{m}\sum_{i=1}^m f_\tau(\alpha_i)
    \;\le\; \frac{\theta_\tau}{2}\,\Var(\valpha).
  \end{equation}

  Exponentiating~\Cref{eq:jensen-two-sided} yields the \emph{pointwise} multiplicative AM--GM control
  (for any fixed $\tau\in[0,1]$):
  \begin{align}
    \exp\!\Bigl(\tfrac{m}{2}\gamma_\tau\,\Var(\valpha)\Bigr)
    \;\le\;
    \frac{\bigl(1-\bar \alpha+\bar \alpha\,\tau\bigr)^m}{\prod_{i=1}^m(1-\alpha_i+\alpha_i\tau)}
    \;\le\;
    \exp\!\Bigl(\tfrac{m}{2}\theta_\tau\,\Var(\valpha)\Bigr),
  \end{align}
  with equality iff $\Var(\valpha)=0$ (or $\tau\in\{0,1\}$). Equivalently, the
  \emph{additive} gap satisfies;
  \begin{align}
    \label{eq:pointwise-add}
    \bigl(1-\bar \alpha+\bar  \alpha \tau\bigr)^m\!\Bigl(1-e^{-\frac{m}{2}\theta_\tau\Var(\valpha)}\Bigr)
    \ge
    \bigl(1-\bar  \alpha+\bar  \alpha \tau\bigr)^m-\!\!\prod_{i=1}^m(1- \alpha_i+ \alpha_i\tau)
    \ge
    \bigl(1-\bar  \alpha+\bar  \alpha \tau\bigr)^m\!\Bigl(1-e^{-\frac{m}{2}\gamma_\tau\Var(\valpha)}\Bigr).
  \end{align}

  Set $\tau=t^{\beta}$ (the theorem’s common exponent) and multiply~\Cref{eq:pointwise-add} by
  $t^{q-1}$, then integrate $t\in[0,1]$.

  Using:
  \[
    I(q;\valpha,\beta)\ \defeq\ \int_{0}^{1}\!\Biggl[\prod_{i=1}^m \bigl(1-\alpha_i+\alpha_i t^{\beta}\bigr)\Biggr] t^{q-1}\,dt,
    \quad
    B_{\mathrm{AMGM}}(q)\ \defeq\ \int_{0}^{1}\!\bigl(1-\bar \alpha+\bar \alpha\,t^{\beta}\bigr)^m t^{q-1}\,dt,
  \]
  we obtain the deterministic two-sided bound:
  \begin{align}
    \label{eq:integrated-gap}
    \underline{\Delta}(q)
    \;\le\; B_{\mathrm{AMGM}}(q)\;-\;I(q;\valpha,\beta) \;\le\; \overline{\Delta}(q),
  \end{align}
  with $\gamma(t)\defeq(1-t^{\beta})^2$, $\theta(t)\defeq\gamma(t)/t^{2\beta}$, $B_{\mathrm{AMGM}}(q)\defeq \gH_\beta(q,\bar \alpha,\beta)$ and:
  \begin{align}
    \label{eq:integrated-gap-def}
    \underline{\Delta}(q)
    \defeq
    \int_{0}^{1}\! t^{q-1}\bigl(1-\bar \alpha+\bar \alpha\,t^{\beta}\bigr)^m
    \Bigl(1-e^{-\frac{m}{2}\,\gamma(t)\,\Var(\valpha)}\Bigr)\,dt,\nonumber \\
    \overline{\Delta}(q)
    \defeq
    \int_{0}^{1}\! t^{q-1}\bigl(1-\bar \alpha+\bar \alpha\,t^{\beta}\bigr)^m
    \Bigl(1-e^{-\frac{m}{2}\,\theta(t)\,\Var(\valpha)}\Bigr)\,dt,\nonumber
  \end{align}

  Using $1-e^{-x}\le x$ gives the simple upper bound:
  \begin{equation}
    \label{eq:integrated-gap-simple}
    0\;\le\;B_{\mathrm{AMGM}}(q)-I(q;\valpha,\beta)
    \;\le\;
    \frac{m}{2}\,\Var(\valpha)\!
    \int_{0}^{1}\! t^{q-1}\bigl(1-\bar \alpha+\bar \alpha\,t^{\beta}\bigr)^m \theta(t)\,dt,
  \end{equation}
  and $1-e^{-x}\ge \tfrac{x}{1+x}$ yields a corresponding explicit lower bound.
  The bounds in~\Cref{eq:integrated-gap} are tight iff $\Var(\valpha)=0$, in which case $B_{\mathrm{AMGM}}(q)=I(q;\valpha,\beta)$.
\end{proof}

\subsection{Proofs for zero-aware gap}\label[appendix]{appendix:zeroaware}
\zerogap*

\begin{proof}
  Fix $t\in[0,1]$ and write $\tau\defeq t^{\beta}\in[0,1]$.
  Let $f_\tau(\alpha)\defeq \log(1-\alpha+\alpha\tau)$, which is concave on $[0,1]$ with
  \[
    -f_\tau''(\alpha)=\frac{(1-\tau)^2}{(1-(1-\tau)\alpha)^2}
    \in\Bigl[(1-\tau)^2,\,(1-\tau)^2/\tau^2\Bigr].
  \]
  Set the zero-aware weights $\lambda_i\defeq \omega_0(\alpha_i)/\Omega$ and denote the
  $\omega_0$–weighted mean and variance by $\bar\alpha^{\omega_0}=\sum_i\lambda_i\alpha_i$ and
  $\Var^{\omega_0}(\valpha)=\sum_i\lambda_i(\alpha_i-\bar\alpha^{\omega_0})^2$ (with the usual
  convention $\Var^{\omega_0}=0$ if $\Omega=0$).

  By weighted Jensen for the concave $f_\tau$:
  \[
    \sum_{i=1}^m \lambda_i f_\tau(\alpha_i)\ \le\ f_\tau(\bar\alpha^{\omega_0}).
  \]
  The standard second-order (weighted) Jensen gap bound gives:
  \[
    f_\tau(\bar\alpha^{\omega_0})-\sum_{i=1}^m \lambda_i f_\tau(\alpha_i)
    \ \le\ \frac{\theta_\tau}{2}\,\Var^{\omega_0}(\valpha),
    \qquad \theta_\tau\defeq\frac{(1-\tau)^2}{\tau^2}.
  \]
  Multiplying by $\Omega$ and exponentiating yields the \emph{pointwise} zero-aware AM--GM control:
  \[
    0\ \le\ \bar Y(t)-Y(t)
    \ \le\ \bar Y(t)\Bigl(1-e^{-\frac{m}{2}\theta_\tau\,\Var^{\omega_0}(\valpha)}\Bigr),
  \]
  where $Y(t)\!=\!\prod_i(1-\alpha_i+\alpha_i t^\beta)$ and $\bar Y(t)\!=\!(1-\bar\alpha+\bar\alpha
    t^\beta)^m$ (note: we keep the envelope centered at the \emph{plain} mean $\bar\alpha$ as in
  ~\Cref{thm:hyperbound}).

  Using $1-e^{-x}\le x$ and integrating against $t^{q-1}$ gives:
  \[
    0\ \le\ B_{\mathrm{AMGM}}(q)-\gI(q,\valpha,\vbeta)
    \ \le\ \frac{m}{2}\,\Var^{\omega_0}(\valpha)\,
    \int_0^1 t^{q-1}\bar Y(t)\,\frac{(1-t^\beta)^2}{t^{2\beta}}\,dt.
  \]
  By differentiating under the integral sign and the binomial identity
  $\sum_{r=0}^{2}\binom{2}{r}(-1)^r t^{r\beta}=(1-t^\beta)^2$, one obtains the identity (derived in
  the main text):
  \[
    \int_0^1 t^{q-1}\bar Y(t)\,\frac{(1-t^\beta)^2}{t^{2\beta}}\,dt
    \;=\;\frac{\partial}{\partial \bar\alpha}\;
    \sum_{r=0}^{2}\binom{2}{r}(-1)^r\,\gH_\beta\!\bigl(q+r\beta;\bar\alpha,m\bigr)
    \;=\;\widetilde{\gA}\bigl(q;\bar\alpha,m\bigr),
  \]
  valid for $q>2\beta$ by Euler’s integral (and for all $q>0$ by analytic continuation). Combining
  the two displays yields
  \[
    B_{\mathrm{AMGM}}(q)-\gI(q,\valpha,\vbeta)
    \ \le\ \frac{m}{2}\,\Var^{\omega_0}(\valpha)\;\widetilde{\gA}\bigl(q;\bar\alpha,m\bigr)
    \ \defeq\ \sA(q,\valpha,m),
  \]
  which is exactly~\Cref{eq:zerogap}. The bound is zero-aware since $\omega_0(0)=0$ removes inactive
  coordinates from $\Var^{\omega_0}$; it is tight when $\Var^{\omega_0}(\valpha)=0$ (i.e., the
  $\omega_0$–weighted dispersion vanishes), in which case $\bar Y(t)\equiv Y(t)$ and equality holds.
\end{proof}

\subsection{Hölder product bound for heterogeneous exponents}
\label[appendix]{app:holder}

Let $\vbeta=(\beta_i)_{i=1}^m$ and take $d$ distinct values $\{b_1,\dots,b_d\}\subset(0,\infty)$,
and partition indices by $S_j\defeq\{i:\beta_i=b_j\}$ with sizes $m_j\defeq |S_j|$ and
$\sum_{j=1}^d m_j=m$. Define the group means:
\[
  \bar\alpha_j \;\defeq\; \frac{1}{m_j}\sum_{i\in S_j}\alpha_i\in[0,1].
\]
Recall the objective integral:
\[
  \gI(q,\valpha,\vbeta)\ \defeq\ \int_0^1 \Biggl[\prod_{i=1}^m\bigl(1-\alpha_i+\alpha_i t^{\beta_i}\bigr)\Biggr]\,t^{q-1}\,dt,
  \qquad q>0.
\]

\begin{lemma}[Hölder–binned envelope]
  \label{lem:holder-bins}
  With the notation above,
  \[
    \gI(q,\valpha,\vbeta)
    \;\le\;
    \prod_{j=1}^d \Bigl[\ \gH_{b_j}\bigl(q;\bar\alpha_j,m\bigr)\ \Bigr]^{\,m_j/m},
  \]
  where $\gH_{\beta}(q;\bar\alpha,m)=\frac{1}{q}\,{}_2F_1(-m,1;\tfrac{q}{\beta}+1;\bar\alpha)$ is the
  common–$\beta$ envelope from~\Cref{thm:hyperbound}.
\end{lemma}

\begin{proof}
  For each group $S_j$ (fixed $t\in[0,1]$):
  \[
    P_j(t)\ \defeq\ \prod_{i\in S_j}\bigl(1-\alpha_i+\alpha_i t^{b_j}\bigr)
    \ \le\ \bigl(1-\bar\alpha_j+\bar\alpha_j t^{b_j}\bigr)^{m_j}
    \quad\text{by AM--GM.}
  \]
  Multiplying over $j$ gives:
  \[
    \prod_{i=1}^m\bigl(1-\alpha_i+\alpha_i t^{\beta_i}\bigr)
    \ \le\
    \prod_{j=1}^d \bigl(1-\bar\alpha_j+\bar\alpha_j t^{b_j}\bigr)^{m_j}.
  \]
  Hence:
  \[
    \gI(q,\valpha,\vbeta)
    \ \le\
    \int_0^1 \prod_{j=1}^d \bigl(1-\bar\alpha_j+\bar\alpha_j t^{b_j}\bigr)^{m_j}\,t^{q-1}\,dt.
  \]

  Let $w_j\defeq m_j/m$ and split $t^{q-1}=\prod_{j=1}^d t^{(q-1)w_j}$. Set:
  \[
    g_j(t)\ \defeq\ \bigl(1-\bar\alpha_j+\bar\alpha_j t^{b_j}\bigr)^{m_j}\,t^{(q-1)w_j}.
  \]
  Choose exponents $p_j\defeq \frac{m}{m_j}>1$, so that $\sum_{j=1}^d \frac{1}{p_j}=\sum_j
    \frac{m_j}{m}=1$. By Hölder’s inequality for products,
  \[
    \int_0^1 \prod_{j=1}^d g_j(t)\,dt
    \ \le\
    \prod_{j=1}^d \Bigl(\int_0^1 |g_j(t)|^{p_j}\,dt\Bigr)^{1/p_j}.
  \]
  But $g_j^{p_j}(t)=\bigl(1-\bar\alpha_j+\bar\alpha_j t^{b_j}\bigr)^{m}\,t^{q-1}$, hence:
  \[
    \int_0^1 \prod_{j=1}^d g_j(t)\,dt
    \ \le\
    \prod_{j=1}^d \Bigl(\int_0^1 \bigl(1-\bar\alpha_j+\bar\alpha_j t^{b_j}\bigr)^{m}\,t^{q-1}\,dt\Bigr)^{m_j/m}.
  \]

  For each $j$, the inner integral equals:
  \[
    \int_0^1 \bigl(1-\bar\alpha_j+\bar\alpha_j t^{b_j}\bigr)^{m}\,t^{q-1}\,dt
    \;=\; \gH_{b_j}\bigl(q;\bar\alpha_j,m\bigr)
    \;=\; \frac{1}{q}\,{}_2F_1\!\Bigl(-m,1;\tfrac{q}{b_j}+1;\bar\alpha_j\Bigr),
  \]
  by the same change-of-variables/Euler-integral used in~\Cref{thm:hyperbound} (valid for $q>0$,
  $b_j>0$). Combining (i)–(iii) yields the stated bound.
\end{proof}

\paragraph{Remarks.}
If exponents are grouped into \emph{bins} with ranges $[b_j^\leftarrow,b_j^\rightarrow]$ rather
than singletons, the same proof holds after replacing $b_j$ by any representative
$b_j^\leftarrow\le \beta_i$ for $i\in S_j$, preserving the upper-bound direction. The bound is a
weighted geometric mean of $d$ hypergeometric envelopes, with weights $m_j/m$, and avoids
collapsing all exponents to a single conservative value.

\paragraph{Temperature–annealed probabilities tighten the zero-aware gap.}
Parameterize the assignment probabilities from logits $Z$ at temperature $\tau>0$:
\[
  p_{i\ell}(\tau)=\mathrm{softmax}\!\left(\frac{Z_{i\ell}}{\tau}\right)\quad\text{(multiclass)}
  \qquad\text{or}\qquad
  p_{i\ell}(\tau)=\sigma\!\left(\frac{Z_{i\ell}}{\tau}\right)\quad\text{(binary)}.
\]
As $\tau\downarrow 0$, $p_{i\ell}(\tau)\to\{0,1\}$ elementwise. Our zero-aware gap in bin $j$ uses
weights $\omega(x)$ (e.g., $\omega(x)=x(1-x)$ or more generally $\omega(x)=x^a$, $a\!\in\![1,2]$),
the weighted mean $\mu=\bar p_{\ell j}^{\omega}$, and dispersion $V=\Var_{\ell j}^{\omega}(p)$
(\Cref{app:final-objective-grad}). Two facts hold:
\begin{enumerate}
  \item \textbf{Vanishing weights at the extremes.} For the choices above, $\omega(0)=\omega(1)=0$ and
        $0\le \omega(x)\le \tfrac{1}{4}$, so for almost-hard assignments $p_{i\ell}(\tau)\in\{0,1\}$
        one has $\Omega_{\ell j}(\tau)=\sum_{r\in S_{\ell j}}\omega\bigl(p_{r\ell}(\tau)\bigr)\ \xrightarrow[\tau\downarrow 0]{}\ 0$
        and $V(\tau)\xrightarrow[\tau\downarrow 0]{}0$.
  \item \textbf{Zero-aware gap collapses.} The (per-bin) gap upper bound
        \[
          \Gamma^{\mathrm{ewa}}_{\ell j}(q)
          =\frac{m_\ell}{2}\,w_{\ell j}\;V\;\widetilde{\gA}_{b_j}\!\bigl(q;\bar p_{\ell j},m_\ell\bigr)
        \]
        vanishes as $\tau\downarrow 0$ because $V(\tau)\to 0$ while $\widetilde{\gA}_{b_j}$ stays bounded
        (finite polynomial in $\bar p$). Hence the total objective’s slack from the AM–GM step is driven to
        zero by temperature annealing.
\end{enumerate}
In contrast, the Hölder envelope terms depend on the \emph{bin means} $\bar p_{\ell j}(\tau)=m_{\ell j}^{-1}\sum_{i\in S_{\ell j}}p_{i\ell}(\tau)$
and thus are insensitive to per-bin \emph{dispersion}. Annealing shrinks only the gap (and any other dispersion-based penalties),
tightening the overall upper bound without altering the envelope’s functional form.

Combining the relaxed envelope and annealing gives the practical surrogate
\[
  \mathcal{U}_{\mathrm{relax}}(P;\tau)
  \;=\;
  \underbrace{\prod_{j=1}^d\!\left[\frac{1}{q}\,{}_2F_1\!\bigl(-m,1;2;\bar p_{\ell j}(\tau)\bigr)\right]^{\!w_{\ell j}}}_{\text{uniform-$c$ Hölder envelope}}
  \;+\;
  \rho\sum_{j}\underbrace{\frac{m_\ell}{2}\,w_{\ell j}\;V_{\ell j}^{\omega}(P(\tau))\;\widetilde{\gA}_{b_j}\!\bigl(q;\bar p_{\ell j}(\tau),m_\ell\bigr)}_{\text{zero-aware gap}},
\]
where $w_{\ell j}=m_{\ell j}/m_\ell$ and $\rho\ge 0$. As $\tau\downarrow 0$, $V_{\ell
      j}^{\omega}(P(\tau))\!\to\!0$ and the gap vanishes, while the envelope is upper-bounded uniformly
by the simple \(c=2\) hypergeometric polynomial in the bin means.

\newpage
\section{Experimental Framework}\label[appendix]{app:experimental}

\subsection{Graph Construction}
Given a dataset $X \in \mathbb{R}^{N \times D}$ consisting of $N$ samples with dimension $D$, we
construct the affinity graph as follows:
\begin{enumerate}
  \item \textbf{$k$-NN Graph:} We compute the $k$-nearest neighbors for each sample based on Euclidean distance, with $k=50$.
  \item \textbf{Adjacency Matrix:} A sparse adjacency matrix $W$ is initialized such that $W_{ij} = 1$ if $j \in \mathcal{N}(i)$ or $i \in \mathcal{N}(j)$.
  \item \textbf{Gaussian Kernel:} The binary edge weights are smoothed using a Gaussian kernel:
        \begin{equation}
          W_{ij} = \exp\left(-\frac{\|x_i - x_j\|^2}{2\sigma_i^2}\right),
        \end{equation}
        where $\sigma_i$ is set to the average distances of neighbors of node $i$.
  \item \textbf{Symmetrization:} We ensure the graph is undirected by updating $W \leftarrow (W + W^\top) / 2$.
\end{enumerate}

\subsection{Probabilistic Assignments and Architecture}
We employ a lightweight neural network $f_\theta$ (implemented as a single linear layer) to map
input features $x_i$ to cluster logits $z_i \in \mathbb{R}^K$, where $K$ is the target number of
clusters. Soft cluster assignments $P \in \mathbb{R}^{N \times K}$ are obtained via the softmax
function:
\begin{equation}
  p_{ik} = \frac{e^{z_{ik}}}{\sum_{j=1}^K e^{z_{ij}}}.
\end{equation}

\subsection{Optimization Objective}
Our method \textbf{H-Cut} objective minimizes an objective that combines a pairwise similarity loss
with a hypergeometric scaling term (to approximate RCut or NCut) and an entropy regularization
term.

\paragraph{Pairwise Similarity Loss.}
For a batch of sampled edges $\mathcal{B} \subset \mathcal{E}$ with weights $w_{ij}$, we minimize
the cross-entropy between connected nodes to encourage smoothness over the graph:
\begin{equation}
  \mathcal{L}_{\text{sim}} = \frac{1}{\sum_{(i,j) \in \mathcal{B}} w_{ij}} \sum_{(i,j) \in \mathcal{B}} w_{ij} \sum_{k=1}^K -p_{ik} \log p_{jk}.
\end{equation}

\paragraph{Hypergeometric Scaling.}
To enforce balanced partitions; acting as a differentiable proxy for the cut volume constraints; we
scale the similarity loss using the Gauss hypergeometric function ${}_2F_1$. We maintain a moving
average of the mean cluster probabilities, denoted $\alpha_k$. The scaling factor $S_k$ for cluster
$k$ is defined as:
\begin{equation}
  S_k = {}_2F_1(-m, b, c, \alpha_k),
\end{equation}
where we set $m=512$, $b=1$, and $c=2$ for H-RCut. This term penalizes clusters that grow disproportionately large.

\paragraph{Entropy Regularization.}
To prevent trivial solutions (collapse to a single cluster), we maximize the entropy of the
marginal cluster distribution $\bar{p}$, where $\bar{p}_k = \frac{1}{|\mathcal{B}|} \sum_{i \in
    \mathcal{B}} p_{ik}$:
\begin{equation}
  \mathcal{L}_{\text{bal}} = -\mathcal{H}(\bar{p}) = \sum_{k=1}^K \bar{p}_k \log \bar{p}_k.
\end{equation}

\paragraph{Total Loss.}
The final objective balances the scaled similarity and the regularization:
\begin{equation}
  \mathcal{L} = \sum_{k=1}^K (\mathcal{L}_{\text{sim}, k} \cdot S_k) + \lambda \mathcal{L}_{\text{bal}}.
\end{equation}
We utilize a gradient mixing strategy to balance the magnitude of gradients between the two terms for stable optimization.

\subsection{Training Algorithm}
The model is trained via Stochastic Gradient Descent (SGD) by sampling edges from the pre-computed
sparse graph. The procedure is summarized below:

\begin{algorithm}[h]
  \caption{H-Cut Training Loop}
  \begin{algorithmic}[1]
    \State \textbf{Input:} Features $X$, Clusters $K$, Pre-computed Edge List $E$
    \State \textbf{Initialize:} Linear model $f_\theta$, Moving average $\alpha \leftarrow \mathbf{1}/K$
    \While{not converged}
    \State Sample batch of edges $(i, j)$ from $E$
    \State Compute assignments $p_i = \text{softmax}(f_\theta(x_i))$, $p_j = \text{softmax}(f_\theta(x_j))$
    \State Compute pairwise similarity $\mathcal{L}_{\text{sim}}$
    \State Update $\alpha$ using batch mean assignments
    \State Compute scaling factors $S_k$
    \State Compute balance penalty $\mathcal{L}_{\text{bal}}$
    \State Update $\theta$ to minimize weighted $\mathcal{L}_{\text{sim}} + \mathcal{L}_{\text{bal}}$
    \EndWhile
    \State \textbf{Output:} Hard assignments $C_i = \arg\max_k p_{ik}$
  \end{algorithmic}
\end{algorithm}

\section{Binning Strategies}\label[appendix]{app:binning}

\begin{itemize}
  \item \textbf{Equal-Frequency Binning:} We sort the vertices by their weights $\beta_i$ and partition them into $d$ bins of equal size $|S_j| \approx n/d$. This approach ensures that each term in the Hölder product contributes equally to the total envelope. However, for power-law degree distributions common in real-world graphs, the final bin often spans a massive range of values (e.g., covering both moderate and hub nodes). This high intra-bin variance causes the representative $\beta_j^\star$ to drastically underestimate the convexity for hub nodes, loosening the bound.

  \item \textbf{Log-Adaptive (K-Means) Binning:} To address the heavy-tailed nature of degree distributions, we apply $K$-Means clustering to the log-transformed weights $x_i = \log(\beta_i)$. This strategy groups vertices based on geometric proximity, ensuring that nodes are binned by their order of magnitude rather than population count. This aligns with the colinearity condition of the Hölder inequality: since the shape of the generating function $t^\beta$ changes most rapidly for small $\beta$ and stabilizes for large $\beta$, clustering in log-space minimizes the shape distortion between the true node functions and the bin-wise proxy, yielding a significantly tighter upper bound.
\end{itemize}

\section{Simulation Details}\label[appendix]{app:binning_sim}

We verify our bounds on a synthetic three-helix dataset with unbalanced clusters ($|C_1|{=}200$,
$|C_2|{=}|C_3|{=}400$) and a 50-NN RBF graph. Soft assignments are parameterized as
$\mP_{i\ell}=\softmax(z_{i\ell}/\tau)$ with random logits, varying $\tau$ from $10^{-2}$ (hard) to
$10^3$ (uniform). The expected cut is estimated via Monte Carlo (MC) with 1000 samples per
temperature.

\begin{figure}[ht]
  \centering
  \begin{subfigure}[b]{0.48\columnwidth}
    \centering
    \includegraphics[width=\textwidth]{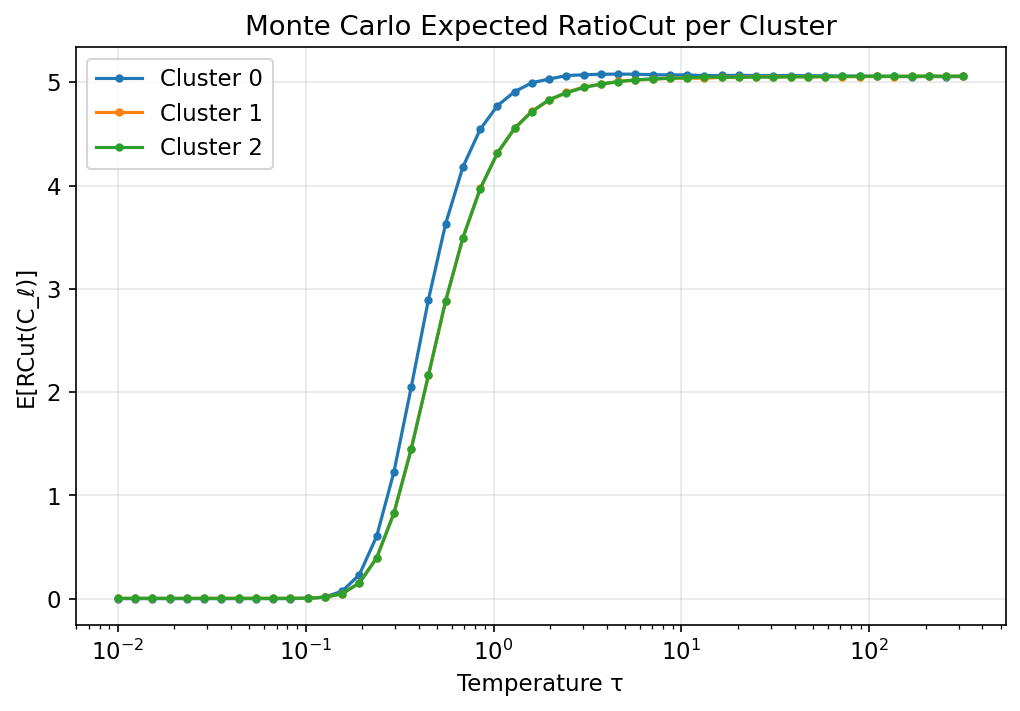}
    \caption{MC expected RCut per cluster}
    \label{fig:mc_rcut}
  \end{subfigure}
  \hfill
  \begin{subfigure}[b]{0.48\columnwidth}
    \centering
    \includegraphics[width=\textwidth]{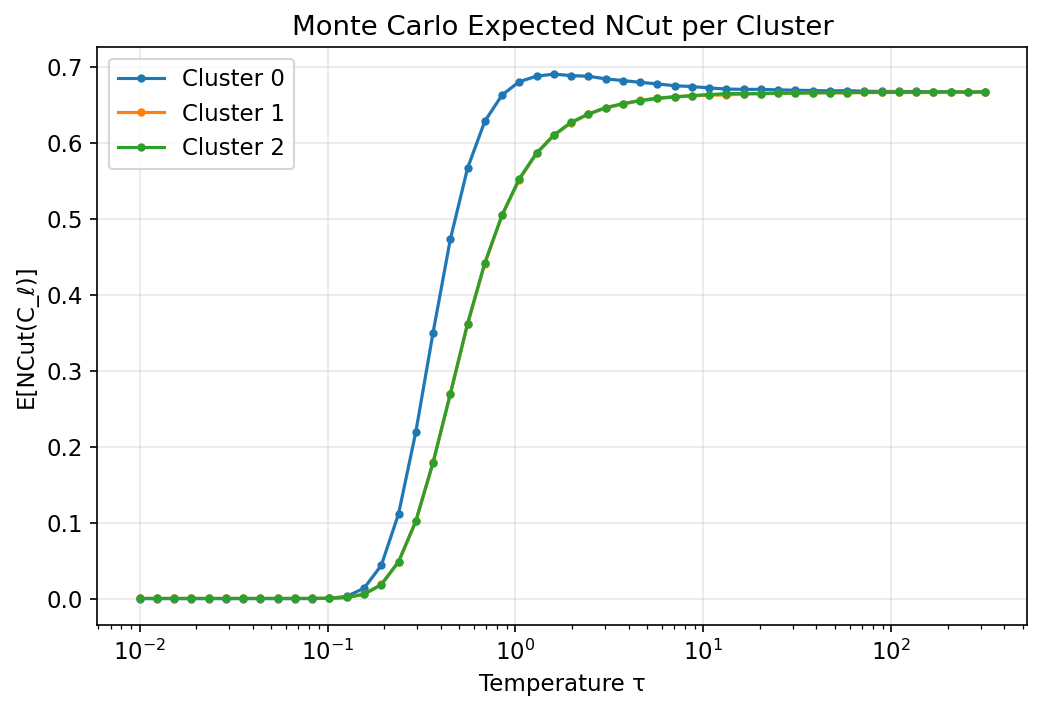}
    \caption{MC expected NCut per cluster}
    \label{fig:mc_ncut}
  \end{subfigure}
  \caption{Monte Carlo expected cut values per cluster as a function of temperature $\tau$.
    The unbalanced cluster ($C_0$, 200 nodes) has higher expected RCut than the balanced ones
    ($C_1,C_2$, 400 nodes each), while NCut normalizes by volume and reduces this gap.}
  \label{fig:mc_simulations}
\end{figure}

\begin{figure}[ht]
  \centering
  \includegraphics[width=0.6\columnwidth]{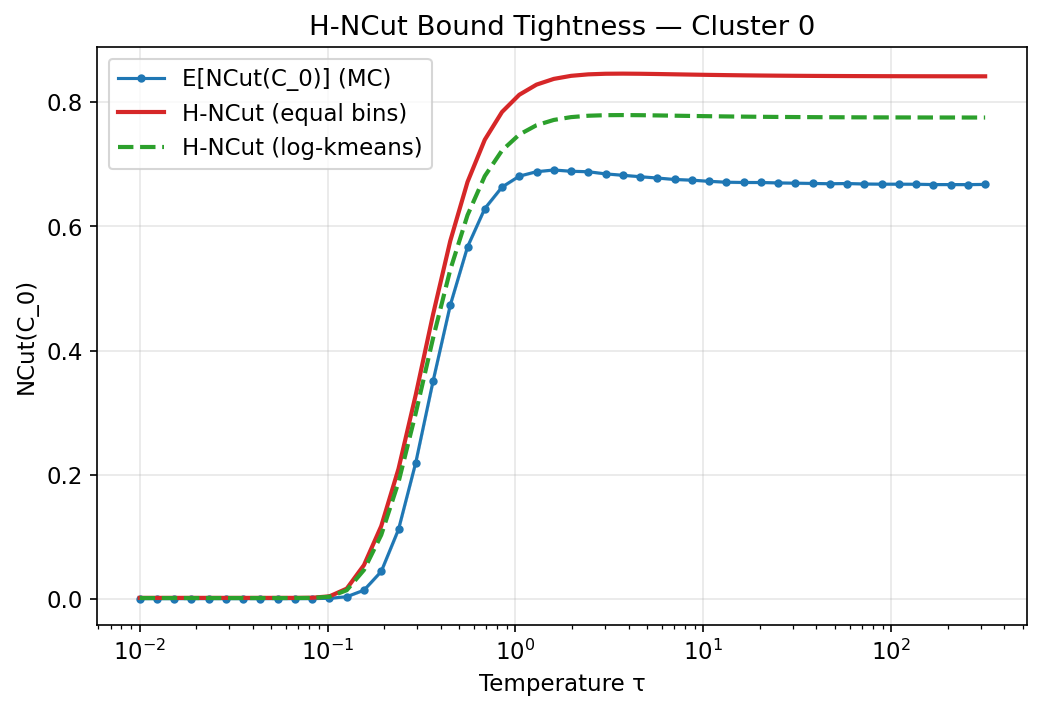}
  \caption{H-NCut bound tightness for the smallest cluster ($C_0$): comparison of
    equal-frequency and log-adaptive binning against the MC estimate. Log-adaptive binning
    tracks the MC curve more closely across all temperatures.}
  \label{fig:ncut_cluster0}
\end{figure}

\begin{figure}[ht]
  \centering
  \includegraphics[width=0.5\columnwidth]{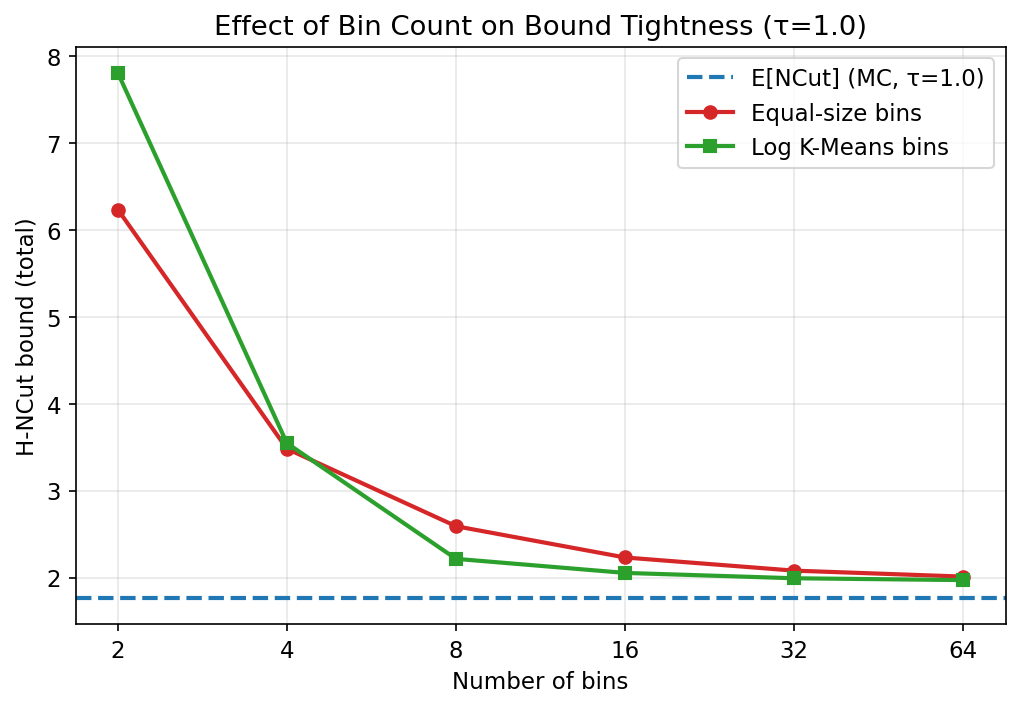}
  \caption{Effect of the number of bins $d$ on H-NCut bound tightness at $\tau{=}1$.
    Log-adaptive binning converges faster and achieves a tighter bound with fewer bins.}
  \label{fig:nbins_sweep}
\end{figure}

\begin{figure}[ht]
  \centering
  \includegraphics[width=0.7\columnwidth]{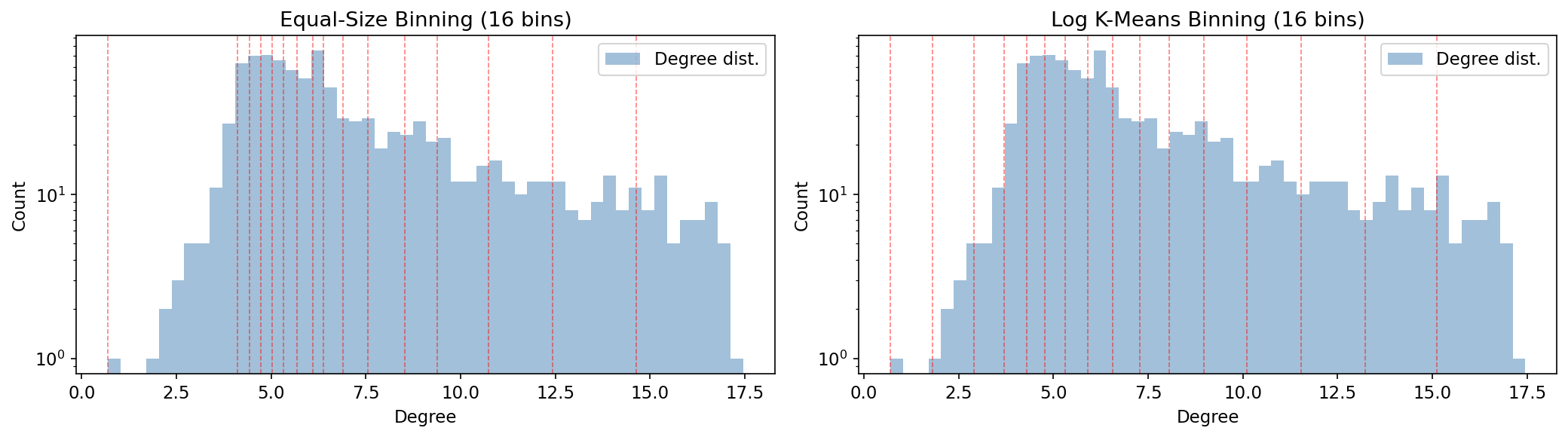}
  \caption{Degree distribution of the helix graph with bin boundaries (red dashed) for
    equal-frequency (left) and log-adaptive K-Means (right) binning with $d{=}16$.
    Log-adaptive binning places more boundaries in the dense region of the distribution.}
  \label{fig:binning_histograms}
\end{figure}

\section{Gradient Mixing Strategy}\label[appendix]{app:gradient_mixing}

A key challenge in optimizing probabilistic graph cuts is balancing the cut objective against a
regularizer that prevents cluster collapse. Without regularization, the optimizer drives small
clusters to zero volume, producing degenerate partitions.

The original PRCut~\citep{prcut} addresses this through its analytical gradient, which includes an
implicit regularization via the $-\text{cut}_\ell / \bar{p}_\ell^2 n$ term. However, this term
vanishes as clusters collapse ($\bar{p}_\ell \to 0$), making the original PRCut prone to empty
clusters in practice; particularly on datasets with many classes ($K \geq 10$).

We introduce a \emph{gradient mixing} strategy that decouples the cut and balance objectives:
\begin{equation}
  \cL = \cL_{\text{cut}} + \cL_{\text{balance}}, \quad
  \cL_{\text{balance}} = -\sum_{\ell=1}^K \bar{p}_\ell \log \bar{p}_\ell,
\end{equation}
where $\bar{p}_\ell = \frac{1}{|\cB|}\sum_{i \in \cB} P_{i\ell}$ is the batch-average
cluster proportion and $\cL_{\text{balance}}$ is the negative entropy, encouraging
uniform cluster sizes. Rather than adding the losses directly (which leads to one
gradient dominating), we normalize each gradient independently before combining:
\begin{equation}
  g_{\theta} = \frac{\nabla_\theta \cL_{\text{cut}}}{\|\nabla_\theta \cL_{\text{cut}}\|}
  + \frac{\nabla_\theta \cL_{\text{balance}}}{\|\nabla_\theta \cL_{\text{balance}}\|}.
\end{equation}
This ensures both objectives contribute equally in gradient direction, regardless of
their relative magnitudes. We refer to PRCut trained with this strategy as PRCut$^*$.

\section{CLIP ViT-L/14 Results}\label[appendix]{app:clip_results}

\begin{table*}[h]
  \centering
  \caption{\textbf{CLIP ViT-L/14 embeddings~\citep{radford21clip}.} CLIP's text-aligned representations yield high $Q$ for coarse categories but struggle with fine-grained distinctions (Flowers, CUB: $Q{=}0.28$). SC = non-parametric Spectral Clustering. Best per row in \textbf{bold}.}
  \label{tab:clipvitL14}
  \resizebox{\textwidth}{!}{
    \begin{tabular}{l|r|c|cccc|cccc|cccc|cccc}
      \toprule
                                               &     &      & \mcfsc        & \mcfpr        & \mcfhr            & \mcfhn                                                                                                                                                                                                                                    \\
      \textbf{Dataset}                         & $K$ & $Q$  & ACC\%         & NMI\%         & RCut $\downarrow$ & NCut $\downarrow$ & ACC\%         & NMI\%         & RCut $\downarrow$ & NCut $\downarrow$ & ACC\%         & NMI\%         & RCut $\downarrow$ & NCut $\downarrow$ & ACC\%         & NMI\%         & RCut $\downarrow$ & NCut $\downarrow$ \\
      \midrule
      CIFAR-10~\citep{cifar}             & 10  & 0.92 & 87.4          & \textbf{89.8} & \textbf{18}       & \textbf{0.40}     & \textbf{89.6} & 87.6          & 37.5              & 0.82              & 83.3          & 82.2          & 40.7              & 0.94              & 83.9          & 82.7          & 40.8              & 0.97              \\
      CIFAR-100~\citep{cifar}            & 100 & 0.61 & \textbf{61.9} & \textbf{71.8} & \textbf{995}      & \textbf{22.6}     & 26.0          & 49.7          & 2307              & 73.2              & 57.6          & 70.8          & 1511              & 35.9              & 57.6          & 70.1          & 1566              & 36.9              \\
      STL-10~\citep{stl10}               & 10  & 0.96 & 95.8          & 95.8          & \textbf{9.0}      & \textbf{0.22}     & \textbf{99.8} & \textbf{99.4} & 9.8               & 0.24              & \textbf{99.8} & \textbf{99.4} & 9.8               & 0.24              & \textbf{99.8} & \textbf{99.4} & 9.8               & 0.24              \\
      EuroSAT~\citep{eurosat}            & 10  & 0.81 & 72.9          & \textbf{73.8} & \textbf{43}       & \textbf{1.0}      & \textbf{78.3} & 70.8          & 68.0              & 1.6               & 76.9          & 73.1          & 64.4              & 1.5               & 76.9          & 73.4          & 62.5              & 1.4               \\
      Imagenette~\citep{imagenette}      & 10  & 0.98 & 99.7          & 99.0          & \textbf{3.2}      & \textbf{0.07}     & 99.7          & 99.2          & 3.6               & 0.08              & \textbf{99.8} & \textbf{99.3} & 3.5               & 0.08              & \textbf{99.8} & \textbf{99.3} & 3.4               & 0.08              \\
      Fashion-MNIST~\citep{fashionmnist} & 10  & 0.76 & \textbf{71.9} & \textbf{73.0} & \textbf{29}       & \textbf{0.66}     & 65.6          & 63.8          & 55.1              & 1.2               & 63.7          & 61.9          & 66.6              & 1.5               & 69.4          & 67.0          & 60.8              & 1.4               \\
      MNIST~\citep{mnist}                & 10  & 0.92 & \textbf{80.9} & \textbf{78.3} & \textbf{19}       & \textbf{0.42}     & 69.4          & 67.1          & 53.4              & 1.2               & 73.5          & 69.7          & 47.1              & 1.1               & 68.0          & 67.5          & 54.9              & 1.3               \\
      Pets~\citep{pets}                  & 37  & 0.60 & 74.5          & 81.2          & \textbf{566}      & \textbf{13.8}     & 81.4          & 85.3          & 567               & 13.9              & 81.3          & 85.4          & 604               & 14.4              & \textbf{83.1} & \textbf{86.0} & 603               & 14.4              \\
      Flowers-102~\citep{flowers}        & 102 & 0.28 & \textbf{81.3} & \textbf{91.0} & 2728              & \textbf{73.7}     & 60.0          & 84.0          & \textbf{2526}     & 80.2              & 70.5          & 85.9          & 3229              & 77.4              & 69.4          & 84.3          & 3081              & 77.5              \\
      Food-101~\citep{food101}           & 101 & 0.76 & \textbf{80.6} & \textbf{85.5} & \textbf{611}      & \textbf{13.6}     & 63.9          & 74.6          & 1259              & 37.8              & 78.3          & 83.9          & 908               & 20.9              & 75.5          & 83.2          & 957               & 21.8              \\
      RESISC-45~\citep{resisc45}         & 45  & 0.77 & 76.2          & 82.9          & \textbf{240}      & \textbf{5.9}      & \textbf{82.3} & \textbf{84.1} & 303               & 7.4               & 77.8          & 82.1          & 324               & 8.0               & 76.8          & 81.9          & 336               & 8.3               \\
      DTD~\citep{dtd}                    & 47  & 0.38 & \textbf{54.3} & 60.9          & \textbf{860}      & \textbf{21.6}     & 52.7          & 61.1          & 938               & 25.5              & 54.1          & 61.1          & 964               & 24.5              & 53.9          & \textbf{61.2} & 995               & 24.2              \\
      GTSRB~\citep{gtsrb}                & 43  & 0.78 & \textbf{68.9} & \textbf{75.7} & \textbf{254}      & \textbf{6.3}      & 64.4          & 68.9          & 469               & 12.3              & 62.5          & 67.9          & 532               & 12.8              & 62.0          & 67.8          & 525               & 12.6              \\
      CUB-200~\citep{cub}                & 200 & 0.28 & \textbf{61.9} & \textbf{81.0} & 5509              & \textbf{140}      & 22.7          & 60.6          & \textbf{4405}     & 179               & 52.3          & 77.5          & 6422              & 151               & 53.3          & 77.9          & 6404              & 150               \\
      FGVC-Aircraft~\citep{aircraft}     & 100 & 0.25 & 37.4          & 58.8          & \textbf{2270}     & \textbf{56.4}     & 36.5          & 59.9          & 2609              & 64.3              & 37.9          & 61.8          & 2742              & 65.9              & \textbf{39.0} & \textbf{62.2} & 2748              & 64.4              \\
      \bottomrule
    \end{tabular}}
\end{table*}

\paragraph{Implementation Details.}
We implement the framework in PyTorch. Optimization is performed using AdamW with a learning rate
of $10^{-4}$ and weight decay of $10^{-4}$. We utilize a large batch size of 8,192 edges on a
single NVIDIA GPU (A100). In our setup, the memory usage does not exceed 3GB of VRAM.


\section{Forward-Backward algorithms}\label[appendix]{appendix:ForwardBackward}

Both envelopes (AM--GM/common--$\beta$ and Hölder/binning) and the zero-aware gap are
differentiable in the assignment parameters $\valpha$ (hence in $\mP$). The backward (pass)
gradients were derived in \S\ref{app:grad-envelope}, \S\ref{app:grad-holder}, and
\S\ref{app:final-objective-grad}. Here we describe the \emph{forward} computation and give robust,
$O(m)$, numerically stable procedures for the truncated hypergeometric terms. Throughout we use:
\[
  \frac{d}{dz}{}_2F_1(a,b;c;z)=\frac{ab}{c}\,{}_2F_1(a+1,b+1;c+1;z),
\]
and the fact that for $a=-m$ (or $-m+1$) the series \emph{truncates} (finite polynomial).

\subsection{Efficient computation of \texorpdfstring{$\hyperg$}{2F1}} \label[appendix]{app:forward-hyperg}

For $a=-m$ and $b=1$, the Gauss hypergeometric reduces to a degree-$m$ polynomial:
\[
  {}_2F_1(-m,1;c;z)
  =\sum_{k=0}^{m}\frac{(-m)_k(1)_k}{(c)_k}\frac{z^k}{k!}
  =\sum_{k=0}^{m}(-1)^k\binom{m}{k}\frac{z^k}{(c)_k},
  \quad z\in[0,1].
\]

Although the series is finite (exact after $k=m$), in practice many tails are negligible. We use an
early-exit rule at index $K$ when:
\[
  |t_{K+1}|<\varepsilon_{\mathrm{rel}}\max\{|S_K|,\delta_{\mathrm{abs}}\},
\]
where $S_K$ is the current partial sum, $\varepsilon_{\mathrm{rel}}$ a relative tolerance, and
$\delta_{\mathrm{abs}}$ a floor for tiny values (e.g., machine epsilon scaled). This is safe
because the remaining $m-K$ terms are alternating and (empirically) rapidly shrinking for
$z\in[0,1]$; for reproducibility one can cap $K\le m$.

At $z=0$ the value is $1$; near $z=1$ we rely on Horner/compensation to manage cancellation. For
large $c$ (e.g., the $c=2$ relaxed envelope of \S\ref{app:holder-relaxed-c2}), coefficients become
very benign: ${}_2F_1(-m,1;2;z)=\sum_{k=0}^m(-1)^k\binom{m}{k}z^k/(k+1)!$.

Applying AM--GM \emph{within} bins $S_j$ of equal $\beta$ (i.e., $\beta_i=b_j$) gives bin-wise
polynomials after replacing $m$ by $m_j$ and $\bar\alpha$ by $\bar\alpha_j$; the product envelope’s
slack is then controlled by within-bin dispersions $\{\Var_j(\valpha)\}_j$.

\subsection{The forward pass (Hölder envelope + derivatives)} \label[appendix]{app:forward-holder}

We now give a concrete forward routine that returns the Hölder envelope
$B_{\mathrm{H\ddot{o}lder}}$ and the two hypergeometric building blocks needed for the backward
pass (the ratio “$H'_j/H_j$” in the gradient). The algorithm evaluates:
\[
  B_{\mathrm{H\ddot{o}lder}}
  =\prod_{j=1}^d\Bigl[\gH_{b_j}(q;\bar\alpha_j,m)\Bigr]^{m_j/m},
  \qquad
  \gH_{b_j}(q;\bar\alpha_j,m)
  =\frac{1}{q}\,{}_2F_1\!\Bigl(-m,1;c_j;\bar\alpha_j\Bigr),
  \quad c_j=\frac{q}{b_j}+1.
\]
We accumulate in the \emph{log domain} to avoid underflow/overflow.

\begin{algorithm}[t]
  \caption{\textsc{HolderBound\&Grad}$(\valpha,\vbeta,q)$}
  \label{alg:holder}
  \begin{algorithmic}[1]
    \State \textbf{Bin} indices by equal $\beta$: obtain $\{(b_j,S_j,m_j,\bar\alpha_j)\}_{j=1}^d$, with $m=\sum_j m_j$.
    \For{$j=1$ to $d$} \Comment{${}_2F_1(-m,1;c_j;\bar\alpha_j)$ via term ratios or Horner}
    \State $c_j\!\gets\!q/b_j+1$;\quad $H_j\!\gets\!1$;\quad $t\!\gets\!1$
    \For{$k=1$ \textbf{to} $m$}
    \State $t\!\gets\!t\cdot \frac{(-m+k-1)}{(c_j+k-1)}\cdot \bar\alpha_j$ \Comment{alternates in sign}
    \State $H_j\!\gets\!H_j+t$ \Comment{use Kahan/Neumaier compensation}
    \If{$|t|<\varepsilon_{\mathrm{rel}}\max(|H_j|,\delta_{\mathrm{abs}})$} \textbf{break} \EndIf
    \EndFor
    \State $B_j^\ast\!\gets\!H_j/q$;\quad $\ell_j\!\gets\!\frac{m_j}{m}\log B_j^\ast$
    \EndFor
    \State $B\!\gets\!\exp\!\Big(\sum_{j=1}^d \ell_j\Big)$ \Comment{$B_{\mathrm{H\ddot{o}lder}}$ in log-sum-exp form}
    \For{$j=1$ to $d$} \Comment{${}_2F_1(-m+1,2;c_j+1;\bar\alpha_j)$ for gradients}
    \State $H'_j\!\gets\!1$;\quad $t\!\gets\!1$
    \For{$k=1$ \textbf{to} $m-1$}
    \State $t\!\gets\!t\cdot \frac{(-m+k)}{(c_j+1+k-1)}\cdot \frac{k+1}{k}\cdot \bar\alpha_j$
    \State $H'_j\!\gets\!H'_j+t$ \Comment{again sign-alternating; compensate}
    \If{$|t|<\varepsilon_{\mathrm{rel}}\max(|H'_j|,\delta_{\mathrm{abs}})$} \textbf{break} \EndIf
    \EndFor
    \EndFor
    \State \textbf{return} $B$ and $\{H_j,H'_j\}_{j=1}^d$ \Comment{used in the backward ratio $H'_j/H_j$}
  \end{algorithmic}
\end{algorithm}

With $H_j={}_2F_1(-m,1;c_j;\bar\alpha_j)$ and $H'_j={}_2F_1(-m+1,2;c_j+1;\bar\alpha_j)$, the
gradient w.r.t.\ an $\alpha_i$ in bin $S_j$ (and zero otherwise) is:
\[
  \frac{\partial B_{\mathrm{H\ddot{o}lder}}}{\partial \alpha_i}
  =-\frac{B_{\mathrm{H\ddot{o}lder}}}{c_j}\,\frac{H'_j}{H_j}\cdot\frac{1}{m_j},
  \qquad c_j=\frac{q}{b_j}+1,
\]
as derived in \S\ref{app:grad-holder}. The same cached $\{H_j,H'_j\}$ also feed the zero-aware gap
gradients in \S\ref{app:final-objective-grad} via $\widetilde{\gA}_{b_j}$ (finite sums of ${}_2F_1$
with shifted parameters).

\paragraph{Complexity and vectorization.}
The forward is $O\!\left(\sum_j m\right)=O(dm)$ scalar ops, embarrassingly parallel over bins. The
backward reuses the same per-bin computations and adds only $O(dm)$ extra ops for $H'_j$ and simple
scalar multiplications.

\paragraph{Forward objective with zero-aware gap.}
The training objective combines the Hölder (or common–$\beta$) envelope with the zero-aware gap
penalty:
\[
  \mathcal{U}(P)
  =\underbrace{\prod_{j=1}^d\Bigl[\gH_{b_j}(q;\bar\alpha_j,m)\Bigr]_{}}_{\text{envelope}}^{m_j/m}
  +\ \rho\sum_{j}\underbrace{\frac{m}{2}\,\frac{m_j}{m}\,V_j^\omega\,\widetilde{\gA}_{b_j}\!\bigl(q;\bar\alpha_j,m\bigr)}_{\text{zero-aware gap}},
\]
where $V_j^\omega$ is the within-bin $\omega$-weighted variance (~\Cref{prop:zeroaware}). Both
terms reuse the same forward hypergeometric blocks; the second depends only on bin means and the
finite differences of the same envelopes.

\subsection{Gradient of the envelope for common \texorpdfstring{$\beta$}{beta}}
\label[appendix]{app:grad-envelope}
Recall the common–$\beta$ envelope from~\Cref{thm:hyperbound}:
\[
  \gH_{\beta}(q;\bar\alpha,m)
  \;=\;
  \frac{1}{q}\;{}_2F_1\!\Bigl(-m,\,1;\,\tfrac{q}{\beta}+1;\,\bar\alpha\Bigr),
  \qquad
  q>0,\ \beta>0,\ \bar\alpha=\tfrac{1}{m}\sum_{i=1}^m \alpha_i.
\]
Since $\gH_\beta$ depends on $\valpha$ only through $\bar\alpha$, the chain rule gives:
\[
  \frac{\partial \gH_\beta}{\partial \alpha_i}
  \;=\;
  \frac{\partial \gH_\beta}{\partial \bar\alpha}\cdot
  \frac{\partial \bar\alpha}{\partial \alpha_i}
  \;=\;
  \frac{1}{m}\,\frac{\partial \gH_\beta}{\partial \bar\alpha},
  \qquad i=1,\dots,m.
\]
Using the standard derivative \( \frac{d}{dz}{}_2F_1(a,b;c;z)=\frac{ab}{c}\;{}_2F_1(a+1,b+1;c+1;z)
\) with $(a,b,c,z)=\bigl(-m,1,\tfrac{q}{\beta}+1,\bar\alpha\bigr)$, we obtain:
\[
  \frac{\partial \gH_\beta}{\partial \bar\alpha}
  =
  \frac{1}{q}\cdot
  \frac{-m}{\tfrac{q}{\beta}+1}\;
  {}_2F_1\!\Bigl(-m+1,\,2;\,\tfrac{q}{\beta}+2;\,\bar\alpha\Bigr),
\]
and hence the per–coordinate gradient:
\[
  \boxed{\;
  \frac{\partial \gH_\beta(q;\bar\alpha,m)}{\partial \alpha_i}
  =
  -\frac{1}{q\bigl(\tfrac{q}{\beta}+1\bigr)}\;
  {}_2F_1\!\Bigl(-m+1,\,2;\,\tfrac{q}{\beta}+2;\,\bar\alpha\Bigr)
  =
  -\frac{\beta}{q(q+\beta)}\;
  {}_2F_1\!\Bigl(-m+1,\,2;\,\tfrac{q}{\beta}+2;\,\bar\alpha\Bigr),
  \quad i=1,\dots,m.}
\]

The gradient is uniform across coordinates because the envelope depends on $\valpha$ only via
$\bar\alpha$. Since $-m$ is a nonpositive integer, ${}_2F_1(-m+1,2;\cdot;\bar\alpha)$ is a
degree-$(m-1)$ polynomial in $\bar\alpha$, enabling stable evaluation via a finite sum or Horner’s
rule.

\subsection{Gradient of the Hölder envelope for heterogeneous \texorpdfstring{$\beta$}{beta}}
\label[appendix]{app:grad-holder}

Recall the Hölder envelope (\Cref{app:holder}): with distinct exponents $\{b_1,\dots,b_d\}$, groups
$S_k\!\defeq\!\{i:\beta_i=b_k\}$, sizes $m_k\!=\!|S_k|$, $m=\sum_k m_k$, and means $\bar\alpha_k
  \!=\! \frac{1}{m_k}\sum_{i\in S_k}\alpha_i$, we defined:
\[
  B_{\mathrm{Holder}}
  \;=\;
  \prod_{k=1}^d \bigl(B_k^\ast\bigr)^{m_k/m},
  \qquad
  B_k^\ast
  \;=\;
  \frac{1}{q}\;{}_2F_1\!\Bigl(-m,\,1;\,c_k;\,\bar\alpha_k\Bigr),
  \quad
  c_k \defeq \frac{q}{b_k}+1,
\]
with $q>0$ and $b_k>0$. We compute the gradient $\partial B_{\mathrm{Holder}}/\partial \alpha_i$
for an index $i\in S_j$ (so $\beta_i=b_j$).

Since $B_{\mathrm{Holder}}$ depends on $\alpha_i$ only through $\bar\alpha_j$,
\[
  \frac{1}{B_{\mathrm{Holder}}}\frac{\partial B_{\mathrm{Holder}}}{\partial \alpha_i}
  \;=\;
  \frac{m_j}{m}\,\frac{1}{B_j^\ast}\,\frac{\partial B_j^\ast}{\partial \alpha_i},
  \qquad
  \frac{\partial B_j^\ast}{\partial \alpha_i}
  \;=\;
  \frac{\partial B_j^\ast}{\partial \bar\alpha_j}\cdot
  \frac{\partial \bar\alpha_j}{\partial \alpha_i}
  \;=\;
  \frac{1}{m_j}\frac{\partial B_j^\ast}{\partial \bar\alpha_j}.
\]
Thus:
\[
  \frac{1}{B_{\mathrm{Holder}}}\frac{\partial B_{\mathrm{Holder}}}{\partial \alpha_i}
  \;=\;
  \frac{1}{m}\,\frac{1}{B_j^\ast}\,\frac{\partial B_j^\ast}{\partial \bar\alpha_j}.
\]

\paragraph{Differentiating the hypergeometric.}
Using $\frac{d}{dz}{}_2F_1(a,b;c;z)=\frac{ab}{c}\,{}_2F_1(a+1,b+1;c+1;z)$ with
$(a,b,c,z)=\bigl(-m,1,c_j,\bar\alpha_j\bigr)$,
\[
  \frac{\partial B_j^\ast}{\partial \bar\alpha_j}
  \;=\;
  \frac{1}{q}\cdot \frac{-m}{c_j}\;
  {}_2F_1\!\Bigl(-m+1,\,2;\,c_j+1;\,\bar\alpha_j\Bigr).
\]
Combining,
\[
  \frac{1}{B_{\mathrm{Holder}}}\frac{\partial B_{\mathrm{Holder}}}{\partial \alpha_i}
  \;=\;
  -\frac{1}{q\,c_j}\,
  \frac{{}_2F_1\!\bigl(-m+1,2;c_j+1;\bar\alpha_j\bigr)}{B_j^\ast}
  \;=\;
  -\frac{1}{c_j}\,
  \frac{{}_2F_1\!\bigl(-m+1,2;c_j+1;\bar\alpha_j\bigr)}{{}_2F_1\!\bigl(-m,1;c_j;\bar\alpha_j\bigr)},
\]
since $B_j^\ast=(1/q)\,{}_2F_1(-m,1;c_j;\bar\alpha_j)$. Multiplying by $B_{\mathrm{Holder}}$ yields
the per–coordinate gradient (identical for all $i\in S_j$, and $0$ for $i\notin S_j$):
\[
  \boxed{\;
    \frac{\partial B_{\mathrm{Holder}}}{\partial \alpha_i}
    \;=\;
    -\frac{B_{\mathrm{Holder}}}{c_j}\,
    \frac{{}_2F_1\!\bigl(-m+1,2;\,c_j+1;\,\bar\alpha_j\bigr)}{{}_2F_1\!\bigl(-m,1;\,c_j;\,\bar\alpha_j\bigr)},
    \qquad
    c_j=\frac{q}{b_j}+1,
    \quad
    i\in S_j.
    \;}
\]

\paragraph{Equivalent forms.}
Let $F_1(z)\!\defeq\!{}_2F_1(-m,1;c_j;z)$ and $F_2(z)\!\defeq\!{}_2F_1(-m+1,2;c_j+1;z)$. By the
derivative identity, \( F_2(z)= -\frac{c_j}{m}\,\frac{d}{dz}F_1(z) \), hence
\[
  \frac{1}{B_{\mathrm{Holder}}}\frac{\partial B_{\mathrm{Holder}}}{\partial \alpha_i}
  \;=\;
  -\frac{1}{c_j}\frac{F_2(\bar\alpha_j)}{F_1(\bar\alpha_j)}
  \;=\;
  \frac{1}{m}\,\frac{d}{dz}\log F_1(z)\Big|_{z=\bar\alpha_j}.
\]
This gives two numerically equivalent implementations:
\begin{align*}
  \text{(ratio form)}\quad          &
  \partial_{\alpha_i}\log B_{\mathrm{Holder}}
  = -\frac{1}{c_j}\,\frac{F_2(\bar\alpha_j)}{F_1(\bar\alpha_j)}, \\
  \text{(log-derivative form)}\quad &
  \partial_{\alpha_i}\log B_{\mathrm{Holder}}
  = \frac{1}{m}\,\frac{d}{dz}\log\! \bigl[{}_2F_1(-m,1;c_j;z)\bigr]\Big|_{z=\bar\alpha_j}.
\end{align*}

Since $-m$ is a nonpositive integer, both hypergeometric terms truncate:
\[
  {}_2F_1(-m,1;c_j;z)
  =\sum_{k=0}^{m}\frac{(-m)_k(1)_k}{(c_j)_k}\frac{z^k}{k!}
  =\sum_{k=0}^{m}(-1)^k\binom{m}{k}\frac{z^k}{(c_j)_k},
\]
\[
  {}_2F_1(-m+1,2;c_j+1;z)
  =\sum_{k=0}^{m-1}\frac{(-m+1)_k(2)_k}{(c_j+1)_k}\frac{z^k}{k!}
  =\sum_{k=0}^{m-1}(-1)^k\,\frac{(m-1)!}{(m-1-k)!}\,\frac{k+1}{(c_j+1)_k}\,z^k.
\]
Thus the ratio in the boxed gradient can be evaluated via stable finite sums (Horner’s rule).

\paragraph{Block structure of the gradient.}
For a fixed bin $j$, all coordinates $i\in S_j$ share the same partial derivative; for $i\notin
  S_j$ the derivative is zero:
\[
  \frac{\partial B_{\mathrm{Holder}}}{\partial \alpha_i}
  =
  \begin{cases}
    -\dfrac{B_{\mathrm{Holder}}}{c_j}\,\dfrac{{}_2F_1(-m+1,2;c_j+1;\bar\alpha_j)}{{}_2F_1(-m,1;c_j;\bar\alpha_j)},
       & i\in S_j,    \\[8pt]
    0, & i\notin S_j.
  \end{cases}
\]

The \emph{log-derivative form} is preferred to avoid overflow/underflow when $m$ is large. Note
that both $F_1$ and $F_2$ are nonnegative on $z\in[0,1]$; the gradient is non-positive (increasing
any $\alpha_i$ weakly decreases the envelope), consistent with the envelope’s monotonicity in
$\bar\alpha_j$. Complexity is $O(d\,m)$ per gradient evaluation using the finite sums across bins;
computation is easy to parallelize over $j$.

\subsection{Gradients of the final objective}
\label[appendix]{app:final-objective-grad}

For cluster $\ell$ and bin index $j$, let $S_{\ell j}$ be the set of vertices assigned to bin $j$
(with common exponent $b_j$), $m_{\ell j}\defeq |S_{\ell j}|$, $m_\ell\defeq\sum_j m_{\ell j}$,
and:
\[
  \bar p_{\ell j}\;\defeq\;\frac{1}{m_{\ell j}}\sum_{r\in S_{\ell j}} p_{r\ell},
  \qquad
  w_{\ell j}\;\defeq\;\frac{m_{\ell j}}{m_\ell}
  \quad(\text{Hölder weight}).
\]
The common–$\beta$ (here $\beta=b_j$) envelope for cluster $\ell$ and bin $j$ is:
\[
  \gH_{b_j}\bigl(q;\bar p_{\ell j},m_\ell\bigr)
  \;=\;
  \frac{1}{q}\,{}_2F_1\!\Bigl(-m_\ell,1;\tfrac{q}{b_j}+1;\bar p_{\ell j}\Bigr),
\]
and the second forward $\beta$–difference and its $\bar p$–derivative are (~\Cref{prop:zeroaware}):
\[
  \gA_{b_j}(q;\bar p_{\ell j},m_\ell)\;\defeq\;
  \sum_{r=0}^{2}\binom{2}{r}(-1)^r\,\gH_{b_j}\!\bigl(q+r b_j;\bar p_{\ell j},m_\ell\bigr),
  \qquad
  \widetilde{\gA}_{b_j}(q;\bar p_{\ell j},m_\ell)\;\defeq\;
  \frac{\partial}{\partial \bar p_{\ell j}}\gA_{b_j}(q;\bar p_{\ell j},m_\ell).
\]

\paragraph{Zero-aware statistics in a bin.}
Fix $i\in S_{\ell j}$ and write
\[
  \omega_i\;\defeq\;\omega(p_{i\ell}),\qquad
  \Omega_{\ell j}\;\defeq\sum_{r\in S_{\ell j}}\omega(p_{r\ell}),\qquad
  S_2\;\defeq\sum_{r\in S_{\ell j}}\omega(p_{r\ell})\,p_{r\ell},
\]
\[
  \mu\;\defeq\;\bar p^{\omega}_{\ell j}\;=\;S_2/\Omega_{\ell j},
  \qquad
  V\;\defeq\;\Var^{\omega}_{\ell j}(p)\;=\;\frac{1}{\Omega_{\ell j}}\sum_{r\in S_{\ell j}}
  \omega(p_{r\ell})\bigl(p_{r\ell}-\mu\bigr)^2,
\]
with the convention $V=0$ if $\Omega_{\ell j}=0$. In the paper we take $\omega(x)=x(1-x)$ so that
$\omega'_i\defeq\frac{d}{dp}\omega(p_{i\ell})=1-2p_{i\ell}$ (zero-aware and symmetric). The
derivatives of the weighted mean and variance are:
\begin{equation}
  \label{eq:mu-var-derivs}
  \frac{\partial \mu}{\partial p_{i\ell}}
  \;=\;\frac{\omega_i+\omega'_i\,(p_{i\ell}-\mu)}{\Omega_{\ell j}},
  \qquad
  \frac{\partial V}{\partial p_{i\ell}}
  \;=\;\frac{1}{\Omega_{\ell j}}
  \Bigl[\,
    \omega'_i\,(p_{i\ell}-\mu)^2
    +2\omega_i\,(p_{i\ell}-\mu)\Bigl(1-\frac{\partial\mu}{\partial p_{i\ell}}\Bigr)
    \,\Bigr]
  -\frac{V}{\Omega_{\ell j}}\omega'_i.
\end{equation}
This is a standard quotient/chain-rule calculation using
$\sum_{r\in S_{\ell j}}\omega(p_{r\ell})(p_{r\ell}-\mu)=0$.

\paragraph{Hypergeometric derivatives needed.}
Let $q_r\defeq q+r b_j$ and $c_r\defeq\tfrac{q_r}{b_j}+1$. Using
$\tfrac{d}{dz}{}_2F_1(a,b;c;z)=\frac{ab}{c}{}_2F_1(a+1,b+1;c+1;z)$,
\begin{align}
  \label{eq:dH-dpbar}
  \frac{\partial}{\partial \bar p_{\ell j}}\gH_{b_j}(q_r;\bar p_{\ell j},m_\ell)
   & =\frac{1}{q_r}\cdot\frac{-m_\ell}{c_r}\;
  {}_2F_1\!\Bigl(-m_\ell+1,\,2;\,c_r+1;\,\bar p_{\ell j}\Bigr), \\
  \label{eq:d2H-dpbar2}
  \frac{\partial^2}{\partial \bar p_{\ell j}^2}\gH_{b_j}(q_r;\bar p_{\ell j},m_\ell)
   & =\frac{1}{q_r}\cdot\frac{-m_\ell}{c_r}\cdot
  \frac{(-m_\ell+1)\cdot 2}{c_r+1}\;
  {}_2F_1\!\Bigl(-m_\ell+2,\,3;\,c_r+2;\,\bar p_{\ell j}\Bigr).
\end{align}
Therefore:
\begin{align}
  \label{eq:dA-tilde}
  \widetilde{\gA}_{b_j}(q;\bar p_{\ell j},m_\ell)
   & =\sum_{r=0}^{2}\binom{2}{r}(-1)^r\,
  \frac{1}{q_r}\cdot\frac{-m_\ell}{c_r}\;
  {}_2F_1\!\Bigl(-m_\ell+1,2;c_r+1;\bar p_{\ell j}\Bigr), \\
  \label{eq:dA-tilde-deriv}
  \frac{\partial}{\partial \bar p_{\ell j}}\widetilde{\gA}_{b_j}(q;\bar p_{\ell j},m_\ell)
   & =\sum_{r=0}^{2}\binom{2}{r}(-1)^r\,
  \frac{1}{q_r}\cdot\frac{-m_\ell}{c_r}\cdot\frac{(-m_\ell+1)\cdot 2}{c_r+1}\;
  {}_2F_1\!\Bigl(-m_\ell+2,3;c_r+2;\bar p_{\ell j}\Bigr).
\end{align}

\paragraph{Zero-aware gap term and its gradient.}
As stated in the paper, our simple zero-aware upper bound for the AM--GM gap in bin $j$ is:
\[
  \Gamma^{\mathrm{ewa}}_{\ell j}(q)\;\defeq\;
  \frac{m_\ell}{2}\,w_{\ell j}\;V\;\widetilde{\gA}_{b_j}\!\bigl(q;\bar p_{\ell j},m_\ell\bigr),
\]
so the per–coordinate gradient for $i\in S_{\ell j}$ is:
\begin{equation}
  \label{eq:grad-gamma-ewa}
  \frac{\partial \Gamma^{\mathrm{ewa}}_{\ell j}}{\partial p_{i\ell}}
  =\frac{m_\ell}{2}\,w_{\ell j}\left[
    \frac{\partial V}{\partial p_{i\ell}}\;\widetilde{\gA}_{b_j}\!\bigl(q;\bar p_{\ell j},m_\ell\bigr)
    \;+\;
    V\;\frac{\partial \widetilde{\gA}_{b_j}}{\partial \bar p_{\ell j}}\cdot
    \frac{\partial \bar p_{\ell j}}{\partial p_{i\ell}}
    \right],
  \qquad
  \frac{\partial \bar p_{\ell j}}{\partial p_{i\ell}}=\frac{1}{m_{\ell j}}.
\end{equation}
Here $\partial V/\partial p_{i\ell}$ and $\partial \mu/\partial p_{i\ell}$ are given by
\Cref{eq:mu-var-derivs}, while $\widetilde{\gA}_{b_j}$ and its derivative are
\Cref{eq:dA-tilde}–\Cref{eq:dA-tilde-deriv}. For $i\notin S_{\ell j}$,
$\partial \Gamma^{\mathrm{ewa}}_{\ell j}/\partial p_{i\ell}=0$.

\paragraph{Envelope term and stick–breaking backward.}
The binned Hölder envelope for cluster $\ell$ (\Cref{app:holder}) is:
\[
  B_{\mathrm{Holder},\ell}\;=\;\prod_{k} \Bigl(\gH_{b_k}(q;\bar p_{\ell k},m_\ell)\Bigr)^{w_{\ell k}},
\]
with per–coordinate gradient (for $i\in S_{\ell j}$):
\[
  \frac{\partial B_{\mathrm{Holder},\ell}}{\partial p_{i\ell}}
  \;=\;
  -\frac{B_{\mathrm{Holder},\ell}}{c_j}\,
  \frac{{}_2F_1\!\bigl(-m_\ell+1,2;c_j+1;\bar p_{\ell j}\bigr)}
  {{}_2F_1\!\bigl(-m_\ell,1;c_j;\bar p_{\ell j}\bigr)}\cdot\frac{1}{m_{\ell j}},
  \qquad c_j=\frac{q}{b_j}+1,
\]
obtained by the same log–diff + chain rule used in~\Cref{app:grad-holder}. (If the outer objective
multiplies the envelope by additional factors; e.g., edge weights $M_{i\ell}(P)$ in the paper—apply
product rule and chain through their own Jacobians.)

Let the final per–cluster contribution be:
\[
  \mathcal{L}_\ell(P)\;=\;U_\ell(P)\;+\;\rho\sum_{j}\Gamma^{\mathrm{ewa}}_{\ell j}(q),
\]
where $U_\ell$ uses the Hölder envelope (possibly multiplied by problem-specific weights), and
$\rho\ge 0$ is the gap regularization. The gradient w.r.t.\ an entry $p_{i\ell}$ is:
\[
  \frac{\partial \mathcal{L}_\ell}{\partial p_{i\ell}}
  \;=\;
  \frac{\partial U_\ell}{\partial p_{i\ell}}
  \;+\;
  \rho\sum_{j:\,i\in S_{\ell j}}
  \frac{\partial \Gamma^{\mathrm{ewa}}_{\ell j}}{\partial p_{i\ell}},
\]
with the explicit pieces given in~\Cref{eq:mu-var-derivs}–\Cref{eq:grad-gamma-ewa}. These feed into
the stick–breaking backward pass exactly as in the main text.

If $\Omega_{\ell j}=0$, set $\mu=0$, $V=0$, and $\partial \mu=\partial V=0$; the bin is inactive
and contributes no gradient. Because $-m_\ell$ is a nonpositive integer, all ${}_2F_1$ terms
truncate to finite polynomials in $\bar p_{\ell j}$, enabling stable Horner evaluation for both
\Cref{eq:dA-tilde} and~\Cref{eq:dA-tilde-deriv}. For $\omega(x)=x^a$ with $a\in[1,2]$, replace
$\omega'_i$ by $a\,p_{i\ell}^{a-1}$ in~\Cref{eq:mu-var-derivs}; the rest of the derivation is
unchanged.

\subsection{A relaxed H\"older envelope}
\label[appendix]{app:holder-relaxed-c2}

\paragraph{Setup.}
Recall the Hölder envelope (\Cref{app:holder}) for heterogeneous exponents:
\[
  B_{\mathrm{Holder}}
  \;=\;
  \prod_{j=1}^d \Bigl(\gH_{b_j}(q;\bar\alpha_j,m)\Bigr)^{m_j/m},
  \qquad
  \gH_{b_j}(q;\bar\alpha_j,m)
  =\frac{1}{q}\,{}_2F_1\!\Bigl(-m,1;\underbrace{c_j}_{=\,q/b_j+1};\bar\alpha_j\Bigr),
\]
where $b_j>0$ is the exponent for bin $j$, $m_j=|S_j|$, $m=\sum_j m_j$, and
$\bar\alpha_j=\frac{1}{m_j}\sum_{i\in S_j}\alpha_i$.

\paragraph{Monotonicity in $c$.}
For $m\!\in\!\mathbb{N}$ and $z\!\in\![0,1]$, the truncated series:
\[
  {}_2F_1(-m,1;c;z)
  =\sum_{k=0}^{m}\frac{(-m)_k(1)_k}{(c)_k}\frac{z^k}{k!}
  =\sum_{k=0}^{m}(-1)^k\binom{m}{k}\frac{z^k}{(c)_k}
\]
has nonnegative terms in absolute value and \emph{each} Pochhammer factor
$(c)_k=c(c+1)\cdots(c+k-1)$ is strictly increasing in $c$. Hence the whole sum is \emph{decreasing}
in $c$:
\[
  c_1\le c_2 \ \Longrightarrow\  {}_2F_1(-m,1;c_1;z)\ \ge\ {}_2F_1(-m,1;c_2;z).
  \tag{$\star$}
  \label{eq:mono-c}
\]

Within a bin $j$, choose a left–endpoint representative $b_j^\leftarrow\le \beta_i$ for $i\in S_j$
(as in~\Cref{app:holder}). Then $c_j^\leftarrow\!=\!q/b_j^\leftarrow+1 \ge q/\beta_i+1$ and, in
particular, if $q\!\ge\!b_j^\leftarrow$ we have $c_j^\leftarrow\!\ge\!2$. Combining the binwise
AM–GM (intra-bin) and Hölder (across bins) steps with the monotonicity~\Cref{eq:mono-c} yields the
\emph{relaxed} envelope:
\[
  \gH_{b_j}(q;\bar\alpha_j,m)
  \;=\;\frac{1}{q}\,{}_2F_1\!\Bigl(-m,1;c_j^\leftarrow;\bar\alpha_j\Bigr)
  \ \le\ \frac{1}{q}\,{}_2F_1\!\Bigl(-m,1;2;\bar\alpha_j\Bigr),
  \quad\text{whenever } c_j^\leftarrow\ge 2.
\]
Therefore:
\[
  \boxed{\;
    B_{\mathrm{Holder}}
    \ \le\
    \underbrace{\prod_{j=1}^d\!\left[\frac{1}{q}\,{}_2F_1\!\bigl(-m,1;2;\bar\alpha_j\bigr)\right]^{\!m_j/m}}_{\displaystyle \
      \defeq\ B_{\mathrm{relax}(c\!=\!2)}}
    \qquad(\text{provided } q\ge b_j^\leftarrow\ \forall j).
    \;}
\]
Intuitively, replacing $c_j$ by the uniform lower value $2$ (the “largest” case
by~\Cref{eq:mono-c}) gives a looser but simpler upper bound. It preserves bin structure through the
\(\bar\alpha_j\)’s and weights \(m_j/m\), but removes the explicit $b_j$–dependence from the
hypergeometric parameter.

\paragraph{Practical simplifications for $c=2$.}
Because $-m$ is a nonpositive integer, ${}_2F_1(-m,1;2;z)$ is a degree-$m$ polynomial in $z$ and
can be evaluated stably by a finite sum (Horner’s rule):
\[
  {}_2F_1(-m,1;2;z)
  =\sum_{k=0}^{m}(-1)^k\,\binom{m}{k}\,\frac{z^k}{(2)_k}
  =\sum_{k=0}^{m}(-1)^k\,\binom{m}{k}\,\frac{z^k}{(k+1)!}.
\]
Thus:
\[
  B_{\mathrm{relax}(c\!=\!2)}
  =\prod_{j=1}^d\left[\frac{1}{q}\sum_{k=0}^{m}(-1)^k\,\binom{m}{k}\,\frac{\bar\alpha_j^{\,k}}{(k+1)!}\right]^{\!m_j/m}.
\]
This form is handy when one wants to precompute per-bin polynomials in $\bar\alpha_j$ independent
of $b_j$.